%% file: main.tex
\documentclass{article}
\usepackage[colorlinks=true,urlcolor=black]{hyperref}
\usepackage{graphicx,subcaption,ragged2e}

\usepackage[preprint]{neurips_2023}

\usepackage[utf8]{inputenc} 
\usepackage[T1]{fontenc}    
\usepackage{hyperref}       
\usepackage{url}            
\usepackage{booktabs}       
\usepackage{amsfonts}       
\usepackage{nicefrac}       
\usepackage{microtype}      
\usepackage[dvipsnames]{xcolor}         
\usepackage{color-edits}    
\usepackage{pifont}
\usepackage{multirow}
\usepackage{wrapfig}
\usepackage{caption} 
\usepackage{floatrow}
\usepackage[frozencache=true,cachedir=minted-cache]{minted}
\usepackage{svg}
\newcommand{\cmark}{\textcolor{green}{\ding{51}}}%
\newcommand{\xmark}{\textcolor{red}{\ding{55}}}%
\addauthor{gn}{magenta}

\hypersetup{
    citecolor = {OliveGreen},
    linkcolor = {BrickRed}
}

\newcommand{\resultsection}[2]{%
    \section{#1 \hspace{2pt} \href{#2}{\raisebox{-2pt}{{\includegraphics[height=.9em]{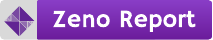}}}}}}

\title{An In-depth Look at Gemini's Language Abilities}
\author{Syeda Nahida Akter$^{*,1}$, Zichun Yu$^{*,1}$, Aashiq Muhamed$^{*,1}$, Tianyue Ou$^{*,1}$, Alex Bäuerle$^{1}$ \\ \textbf{Ángel Alexander Cabrera$^{1}$, Krish Dholakia$^{2}$, Chenyan Xiong$^{1}$, Graham Neubig$^{1}$} \\ $^{1}$Carnegie Mellon University, $^{2}$BerriAI}
\date{December 2023}

\begin{document}

\maketitle
\def\thefootnote{*}\footnotetext{Lead authors. Individual author contributions are listed in \autoref{sec:author_contributions}.}\def\thefootnote{\arabic{footnote}}

\begin{abstract}
  The recently released Google Gemini class of models are the first to comprehensively report results that rival the OpenAI GPT series across a wide variety of tasks.
  In this paper, we do an in-depth exploration of Gemini's language abilities, making two contributions.
  First, we provide a third-party, objective comparison of the abilities of the OpenAI GPT and Google Gemini models with reproducible code and fully transparent results.
  Second, we take a closer look at the results, identifying areas where one of the two model classes excels.
  We perform this analysis over 10 datasets testing a variety of language abilities, including reasoning, answering knowledge-based questions, solving math problems, translating between languages, generating code, and acting as instruction-following agents.
  From this analysis, we find that Gemini Pro achieves accuracy that is close but slightly inferior to the corresponding GPT 3.5 Turbo on all English-language tasks that we benchmarked, but find that Gemini Pro excels in translation into other languages for the languages that it supports.
  We further provide explanations for some of the under-performing tasks, including failures in mathematical reasoning with many digits, sensitivity to multiple-choice answer ordering, and others.
  We also identify areas where Gemini Pro demonstrates comparably high performance, such as handling longer and more complex reasoning chains.
  Code and data for reproduction can be found at  \texttt{\url{https://github.com/neulab/gemini-benchmark}}
\end{abstract}

\input{sections/01_Intro}
\input{sections/02_ExpProcedure}

\input{sections/03_Knowledge-based_QA}
\input{sections/04_General_Reasoning}
\input{sections/05_math}

\input{sections/06_Code}

\input{sections/07_MT}
\input{sections/08_WebAgent}

\input{sections/09_Conclusion}

\section*{Acknowledgements}

We greatly appreciate the help of Zhiruo Wang in handling the ODEX dataset, and Shuyan Zhou with high-level guidance on the WebArena experiments.
We also are very grateful to the Gemini team who, based on an earlier version of this draft, provided significant help in attempting to reproduce the numbers in their paper.
We also thank those who provided comments on our earlier paper draft on social media, including Arthur Mensch, who noted that we had inadvertently used a non-official third-party version of the Mixtral model in comparisons, and those who pointed out an inaccuracy in description of the Mixtral model's mixture of experts mechanism.

\bibliographystyle{plainnat}
\bibliography{references}

\appendix

\input{sections/A01_contributions}
\input{sections/A02_prompt_details}
\input{sections/A03_mt_additional}

\end{document}

%% file: sections/01_Intro.tex
\section{Introduction}

Gemini is the most recent in a series of large language models released by Google DeepMind \citep{gemini23gemini}.
It is notable in particular because the results reported by the Gemini team are the first to rival the OpenAI GPT model series \citep{brown2020language} across a wide variety of tasks.
Specifically, Gemini's ``Ultra'' version reportedly outperforms GPT-4 on a wide variety of tasks, while Gemini's ``Pro'' version is reportedly comparable to GPT-3.5 \citep{gemini23gemini}.
Despite the potential impact of these results, the exact evaluation details and model predictions have not been released, limiting the ability to reproduce, inspect, and analyze the results and their implications in detail.

In this paper, we conduct an in-depth exploration of Gemini's language understanding and generation abilities, with two goals:
\begin{enumerate}
\item We aim to provide a \textbf{third-party, objective comparison} of the abilities of the OpenAI GPT and Google Gemini model classes with reproducible code and fully transparent results.
\item We aim to take an \textbf{in-depth look into the results}, identifying areas where one of the two model classes excels.
\end{enumerate}
Furthermore, we also perform a limited comparison with the recently released Mixtral model, as a point of reference for a best-in-class open source model \citep{mistralai_2023_mixtral}.

We perform this analysis over 10 datasets, testing a variety of text understanding and generation capabilities, including the models' abilities to answer knowledge-based questions (MMLU; \citet{hendrycks2020measuring}), perform reasoning (BigBenchHard; \citet{suzgun2022challenging}), answer mathematics questions (e.g.~GSM8K; \citet{cobbe2021training}), translate between languages (e.g.~FLORES; \citet{goyal2022flores}), generate code (e.g.~HumanEval; \citet{chen2021evaluating}), and act as an instruction-following agent (WebArena; \citet{zhou2023webarena}).%
\footnote{Note that Gemini is a multi-modal model, but for this examination, we only focus on Gemini's language understanding, generation, and translation abilities.}

A summary of our main results can be found in \autoref{tab:main_results}.
In sum, we found that across all tasks, as of this writing (\today), \textbf{Gemini's Pro model achieved comparable but slightly inferior accuracy compared to the current version of OpenAI's GPT 3.5 Turbo for all English tasks, but superior ability to translate into other languages that it supports.}
Mixtral was competitive with the Gemini and GPT models for Knowledge-based QA and Mathematics tasks, but fell short in more complex tasks.
In the following sections, we will detail our experimental methodology (\autoref{sec:experimental_setup}) and then perform an in-depth description and analysis of the results on each task.
Each analysis is accompanied by an online results browser using Zeno \citep{cabrera2023zeno},\footnote{https://zenoml.com} which can be accessed through the \includegraphics[height=.9em]{figures/zeno-report.pdf} images in this PDF.
All results and code for reproduction can be found at \texttt{\url{https://github.com/neulab/gemini-benchmark}}.

\input{tables/main_res}

%% file: tables/main_res.tex
\begin{table}[t]
    \centering
    \resizebox{\textwidth}{!}{%
    \begin{tabular}{lcccccc} 
        \toprule
        & \multicolumn{4}{c}{\textbf{Model}} \\
        \cmidrule{2-6}
        \textbf{Task} & \textbf{Dataset} & Gemini Pro & GPT 3.5 Turbo & GPT 4 Turbo & Mixtral \\
        \midrule
        \multirow{2}{*}{Knowledge-based QA} & MMLU (5-shot) & 65.22 &  67.75 & \textbf{80.48} & \underline{68.81} \\
        & MMLU (CoT) & 62.09 &  \underline{70.07} &  \textbf{78.95} & 59.57 \\
        \midrule
        Reasoning & BIG-Bench-Hard & 67.53 & \underline{71.02} & \textbf{83.90} & 60.76 \\
        \midrule
        \multirow{4}{*}{Mathematics} & GSM8K & 76.42 & \underline{78.01} &  \textbf{92.72} & 71.65 \\
        & SVAMP & 81.10 & \underline{82.30} & \textbf{92.60} & 81.60 \\
        & ASDIV & 85.31 & \underline{89.07} & \textbf{92.75} & 83.16 \\
        & MAWPS & 96.50 & \underline{98.00} & \textbf{98.67} & 96.00 \\
        \midrule
        \multirow{2}{*}{Code Generation} & HumanEval & 59.76 & \underline{74.39} & \textbf{76.83} & 45.12 \\
        & ODEX & 39.86 & \textbf{52.62} & \underline{45.79} & 40.55 \\
        \midrule
        \multirow{2}{*}{Machine Translation}
         & FLORES (5-shot) Unblocked & \underline{53.31}  & 52.43 & \textbf{54.00} & 40.97 \\
         & FLORES (5-shot) All & 21.68  & \underline{40.00} & \textbf{48.24} & 30.27 \\
        \midrule
        \multirow{1}{*}{Web Agents} & WebArena & 7.12 & \underline{8.87} & \textbf{14.90} & 1.39 \\
        \bottomrule
    \end{tabular}
    }
    \vspace{5pt}
    \caption{Main results of our benchmarking. The best model is listed in bold, and the second best model is underlined.}
    \label{tab:main_results}
\end{table}

%% file: sections/02_ExpProcedure.tex
\section{Experimental Setup}
\label{sec:experimental_setup}
Before discussing evaluation results and findings, this section describes our experiment configurations, including models tested, model querying details, and evaluation procedures.

\subsection{Models Tested}

In this work, we compare 4 models.

\paragraph{Gemini Pro} is the second largest model in the Gemini Series, next to the largest Gemini Ultra.%
\footnote{Gemini Ultra is not yet publicly available, and thus we do not test it in the current version of this paper.}
The model is based on the Transformer \citep{vaswani2017attention} architecture and was trained multimodally over videos, text, and images.
The number of parameters and size of training data are not disclosed.
In the original Google paper on Gemini, it was reported to achieve similar performance to GPT 3.5 Turbo.

\paragraph{GPT 3.5 Turbo} is the second most capable text model served by OpenAI, part of the GPT-3 series \citep{brown2020language}.
The model has been instruction tuned and trained using reinforcement learning from human feedback \citep{ouyang2022training}, but was trained solely on text.
Similarly, model size and precise training details are not disclosed.

\paragraph{GPT 4 Turbo} is the second generation of the GPT-4 \citep{openai2023gpt4} family, a family of models trained multimodally.
The turbo version is moderately cheaper than the original GPT-4 model (making it more conducive to benchmarking) and similarly lacks detail of the actual training algorithms, data, or parameter size.

\paragraph{Mixtral} in contrast, is an open-source mixture-of-experts model, where each feedforward block picks from a set of 8 distinct groups of parameters and uses two to process the token \citep{mistralai_2023_mixtral}.
It has been reported to achieve comparable accuracy to GPT 3.5 Turbo on several tasks, including some examined in this paper. We use the \texttt{mistralai/Mixtral-8x7b-Instruct-v0.1} version of the model.

\input{tables/price}

\subsection{Model Querying Details}
\label{sec:model_querying_details}

All models were queried through the unified interface provided by LiteLLM\footnote{\url{https://litellm.ai/}} between December 11-22, 2023.
Gemini was queried through Google Vertex AI, OpenAI models through the OpenAI API, and Mixtral through the API provided by Together.%
\footnote{\url{https://cloud.google.com/vertex-ai/docs} \url{https://openai.com/api} \url{https://docs.together.ai/docs}}
For reference, we also list the current pricing of each model through these APIs for 1M tokens in \autoref{tab:pricing}, which provides an approximate measure of how efficiently the models can be run.

It is also notable that in some cases Gemini Pro, by default, has safety features\footnote{\url{https://cloud.google.com/vertex-ai/docs/generative-ai/multimodal/configure-safety-attributes}} that block some questions, particularly in the case of potentially illegal or sensitive material.
For analysis in this paper, we disabled these safety settings, but in some cases discuss the effect on measured accuracy (contrasting with \texttt{gemini-pro-filtered}, a model with these safety filters enabled).

\subsection{Evaluation Procedure}

To perform a fair comparison between the models, we re-ran experiments with all models using \emph{exactly the same prompts and evaluation protocol for all evaluated models}.
We make this decision to ensure that all models are compared on exactly the same footing, in contrast to previous papers where these settings may differ.
In general, we tried to follow both prompts and evaluators from standard repositories, either those officially released by the datasets themselves, or from the Eleuther evaluation harness \citep{eval-harness}.
We also personally communicated with the Gemini team, and in some cases followed their recommended prompts for evaluating the models, in the cases where these prompts provided uniformly superior to the standard prompts over all evaluated model classes.
These prompts generally consist of a query, input, and few-shot examples, sometimes including chain-of-thought reasoning \citep{wei2022chain}.
In some cases, we found it necessary to make small changes from standard practice to stably evaluate all models under consideration; all such deviations are noted below and implemented in the companion code repository.

%% file: tables/price.tex
\begin{wraptable}{r}{0pt}
\vspace{-5mm}
\begin{tabular}{r|rr}
\toprule
Language Model & Input & Output \\
\hline
    Gemini Pro & \$1.00 & \$2.00  \\
 GPT 3.5 Turbo & \$1.00 & \$2.00 \\
   GPT 4 Turbo & \$10.00 & \$30.00 \\
       Mixtral & \$0.60 & \$0.60 \\
\bottomrule
\end{tabular}
\caption{
Pricing per 1M tokens. Gemini Pro charges by character; we multiply by 4, a rule-of-thumb average of characters per English token \citep{raf2023whataretokens}.
\label{tab:pricing}
}
\end{wraptable}

%% file: sections/03_Knowledge-based_QA.tex
\resultsection{Knowledge-based QA}{https://hub.zenoml.com/report/2674/Gemini\%20MMLU}
\label{sec:mmlu}

In this category, we focus on 57 knowledge-based multiple-choice question-answering tasks from MMLU \citep{hendrycks2020measuring}, which span topics across STEM, the humanities, the social sciences, and more. MMLU has been widely used as a holistic evaluation of LLMs' knowledge-based capabilities. There are 14,042 test samples in total.

\subsection{Experimental Details}

\paragraph{Generation Parameters}
We examine two popular evaluation methods in this task, including the standard 5-shot prompts from \citet{hendrycks2020measuring} and 5-shot chain-of-thought prompts from chain-of-thought-hub\footnote{\url{https://github.com/FranxYao/chain-of-thought-hub}} \citep{fu2023chain} with a prefix of ``Let's think step by step.'' \citep{takeshilarge2022}. Note that we opt not to sample multiple responses and perform self-consistency based reranking \citep{wang2022self} as done by \citet{gemini23gemini}, as this significantly increases cost and may not be feasible in many scenarios.
We generate via greedy search with a temperature of 0.

\paragraph{Evaluation}
For the standard prompting, we directly take the first character generated by models as their answer since this is what the 5-shot prompts imply.
Sometimes, the model may not follow this format and output the answer elsewhere.
We treat examples like this as incorrect (and elaborate more on the effect of this in the following section).
For the chain-of-thought prompting, we perform answer extraction from the model's response and set the default answer as ``C'' if no answer can be extracted, as is done in chain-of-thought-hub.

\subsection{Results and Analysis}


\input{charts/mmlu_acc_ratio}

In this section, we compare and analyze the overall performance, performance by sub-tasks, and performance by output length on MMLU.

First, from the overall results shown in \autoref{fig:mmlu_overall_accuracy}, we can see that Gemini Pro achieves an accuracy close to, but slightly lower than that of GPT 3.5 Turbo, and much lower than that of GPT 4 Turbo.
MMLU is the strongest task for the Mixtral model, with it besting both Gemini Pro and GPT 3.5 Turbo.
We saw some degradation of performance using chain-of-thought prompting, likely because MMLU is mostly a knowledge-based question answering task that may not benefit significantly from reasoning-oriented prompts.%
\footnote{
Note that our evaluation numbers are slightly lower than those reported in the Gemini \citep{gemini23gemini} and Mixtral \citep{mistralai_2023_mixtral} technical reports.
We attribute this to sensitivity to prompts -- other prompts may achieve somewhat higher accuracy overall.
}


Based on this overall result, we next dive a bit deeper.
One first notable point is that all questions in MMLU are multiple-choice with 4 potential answers ordered A through D.
In \autoref{fig:mmlu_multiple_choice_answer_ratio}, we show the ratio of the number of times each model selects each multiple choice answer.
From this figure, we can see that Gemini has a very skewed label distribution, biased towards selecting the final choice of ``D'', which contrasts to the result of the other models, which are more balanced.
This may indicate that Gemini has not been heavily instruction-tuned towards solving multiple-choice questions, which can cause models to be biased with respect to answer ordering \citep{pezeshkpour2023large,zheng2023large,tjuatja2023llms}.

\begin{figure}
\begin{subfigure}[t]{0.392\textwidth}
    \includegraphics[width=\textwidth]{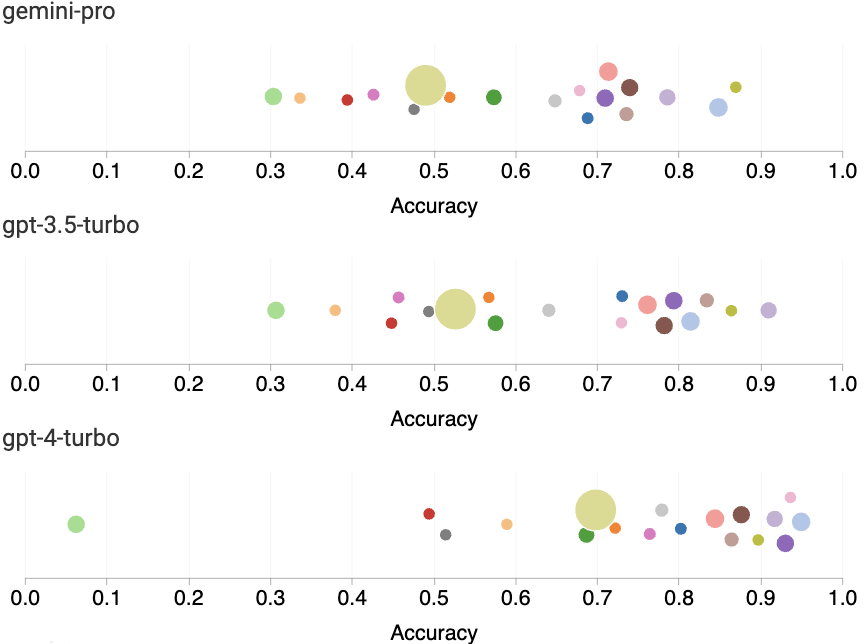}
\end{subfigure}\hspace{\fill} 
\begin{subfigure}[t]{0.58\textwidth}
    \includegraphics[width=\linewidth]{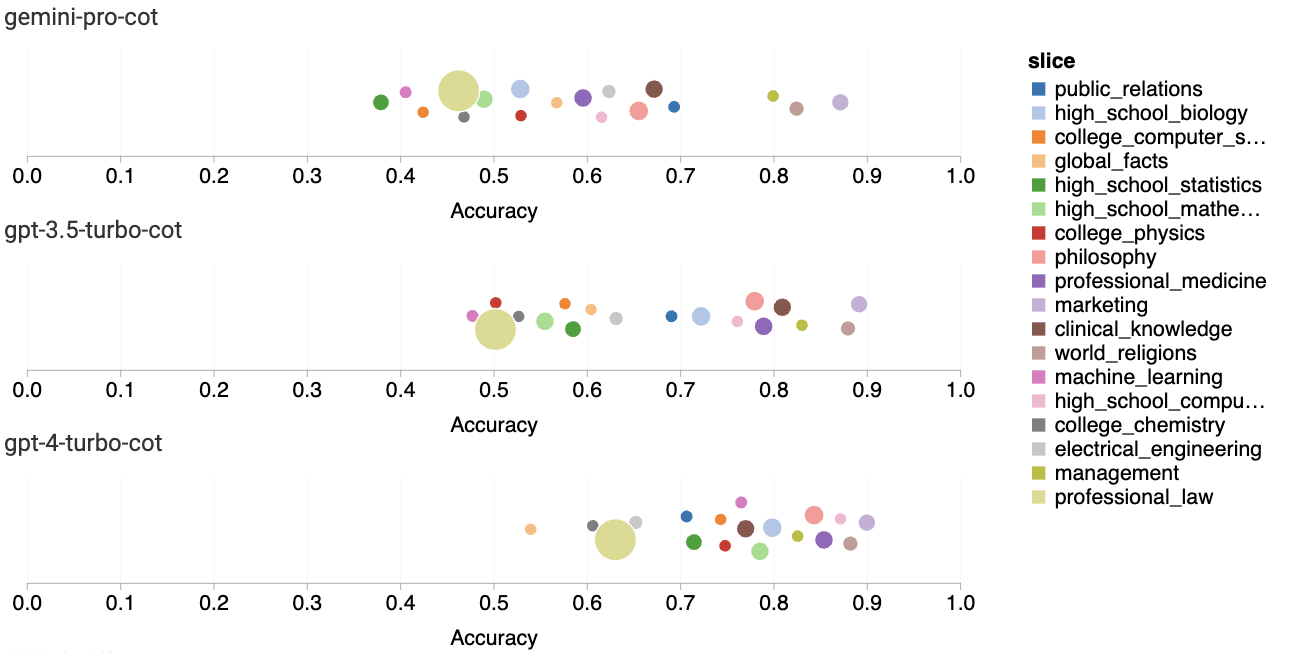}
\end{subfigure}

\caption{Accuracy by each subtask on MMLU}
\label{fig:mmlu_accuracy_by_task}
\end{figure}

Next, we examine each subtask's performance. \autoref{fig:mmlu_accuracy_by_task} illustrates each model's performance on selected representative tasks. We notice that Gemini Pro underperforms on most tasks compared to GPT 3.5 Turbo.
Chain-of-thought prompting decreases the variance across the subtasks.

\begin{figure}
\begin{subfigure}[t]{0.47\textwidth}
    \includegraphics[trim=0 20 0 0, clip, width=\textwidth]{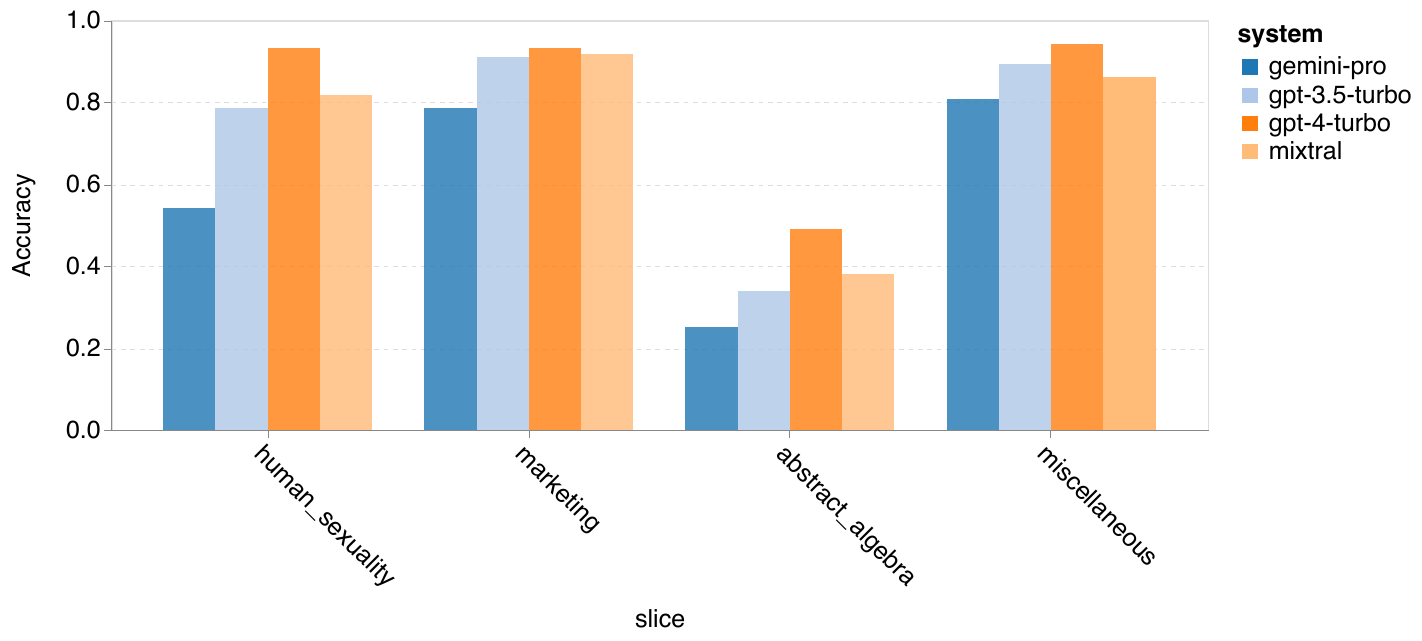}
    \caption{Top-4 tasks where GPT 3.5 wins over Gemini Pro}
\end{subfigure}\hspace{\fill} 
\begin{subfigure}[t]{0.47\textwidth}
    \includegraphics[trim=0 20 0 0, clip, width=\linewidth]{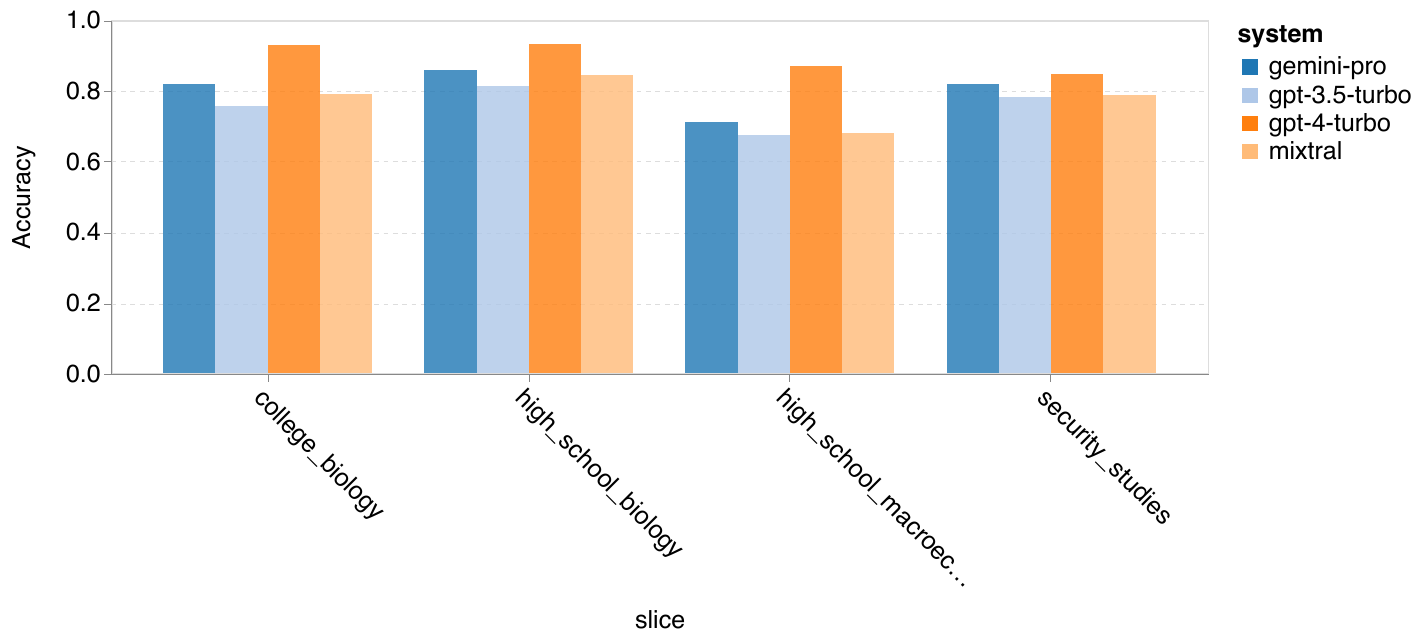}
    \caption{Tasks where Gemini Pro wins over GPT 3.5}
\end{subfigure}

\caption{Tasks where Gemini Pro and GPT 3.5 prevail on MMLU}
\label{fig:mmlu_win_task}
\end{figure}

Further, we dig deeper into the tasks where Gemini Pro underperforms/outperforms GPT 3.5 Turbo the most. From \autoref{fig:mmlu_win_task}, we can observe that Gemini Pro falls behind GPT 3.5 Turbo on \texttt{human\_sexuality}, \texttt{marketing} \texttt{abstract\_algebra}, and \texttt{miscellaneous}.
In contrast, Gemini Pro excels at both \texttt{college\_biology} and \texttt{high\_school\_biology}, as well as \texttt{high\_school\_macroeconomics} and \texttt{security\_studies}.

One important thing to note is that, as previously mentioned in \autoref{sec:model_querying_details}, Gemini Pro's safety features can have a significant effect on overall performance.
All results reported above are Gemini Pro with safety filtering \emph{disabled}, but when the features are enabled the response rate and corresponding performance can drop.
In most MMLU sub-tasks, the API response rate was greater than 95\%, but two had notably low response rates: \texttt{moral\_scenarios} at 85\% and \texttt{human\_sexuality} at 28\%.

\begin{figure}[th]
\begin{subfigure}[t]{0.49\textwidth}
    \includegraphics[trim=0 20 0 0, clip, width=\textwidth]{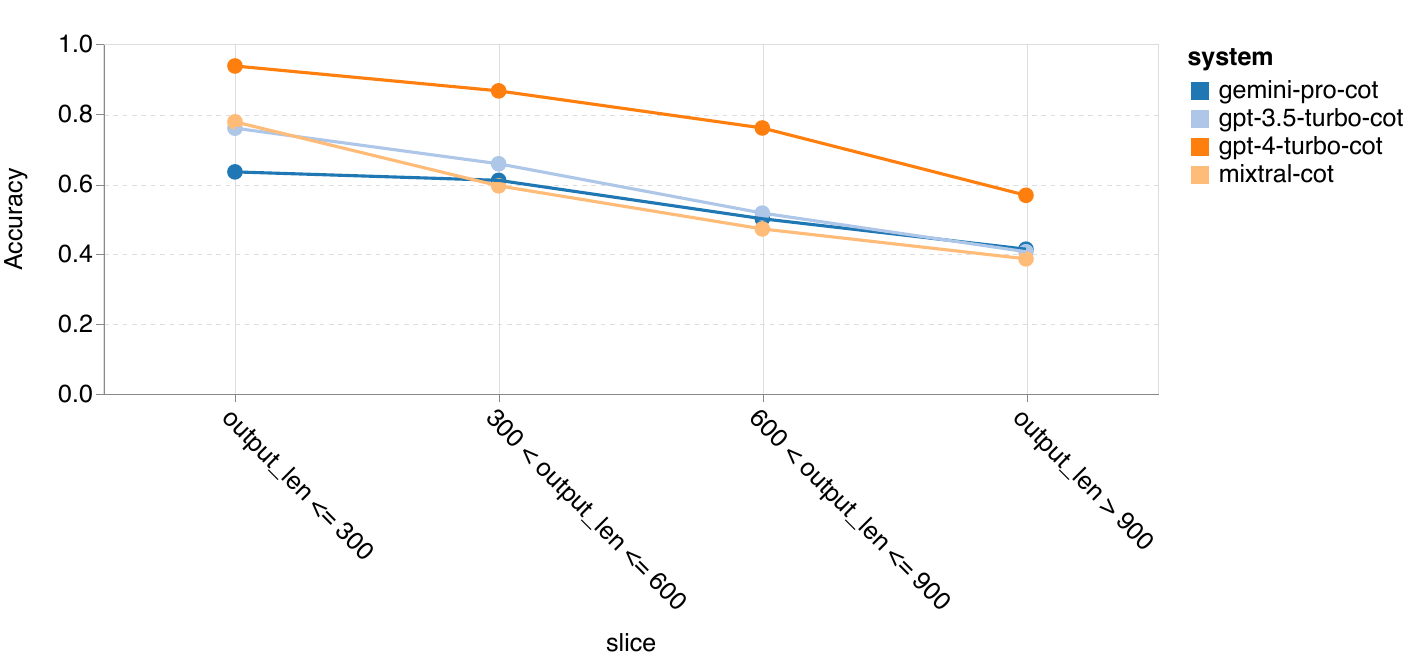}
    \caption{Accuracy by output length}
\end{subfigure}\hspace{\fill} 
~
\begin{subfigure}[t]{0.49\textwidth}
    \includegraphics[trim=0 20 0 0, clip, width=\linewidth]{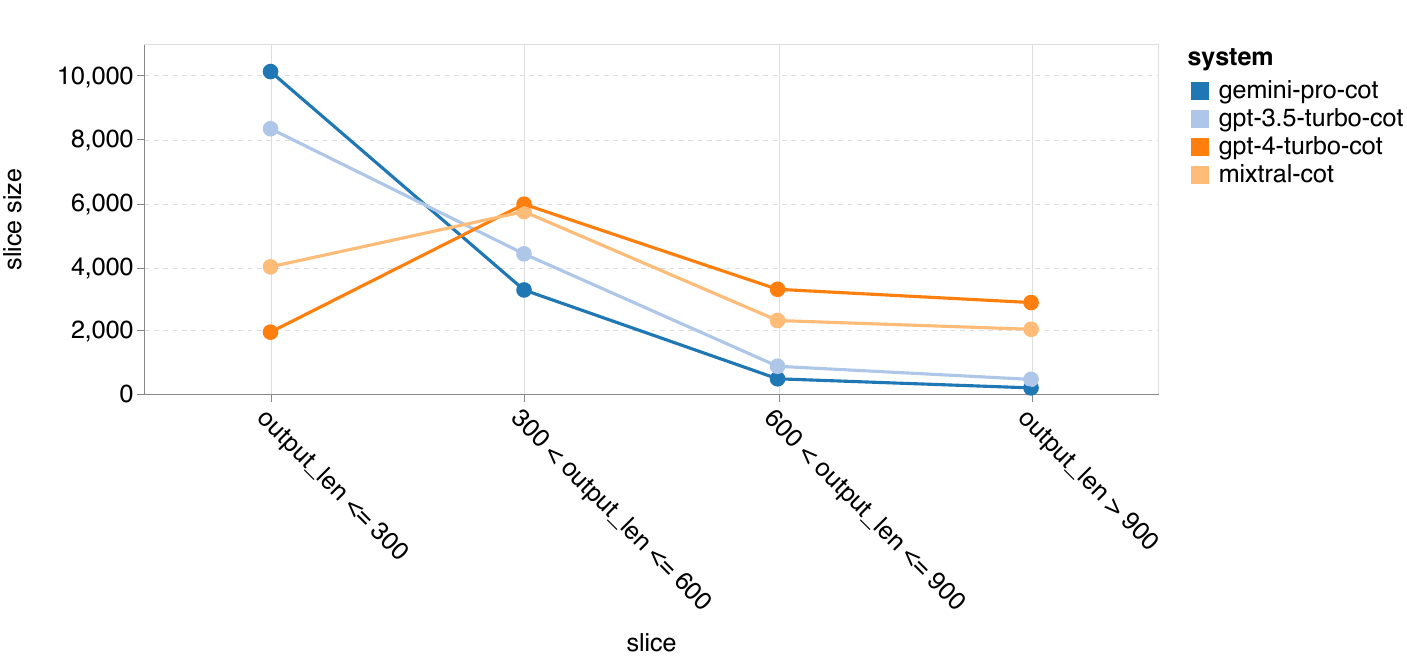}
    \caption{Output length distribution}
\end{subfigure}

\caption{Analysis of output length on MMLU}
\label{fig:mmlu_output_length}
\end{figure}

Finally, we analyze how the output length in the chain-of-thought prompting affects the model performance in \autoref{fig:mmlu_output_length}. Generally, a stronger model tends to perform more complex reasoning and thus outputs a longer response. One of the noteworthy advantages of Gemini Pro is that its accuracy is less influenced by the output length compared to the two counterparts. It even outperforms GPT 3.5 when the output length is over 900. However, it also can be seen that Gemini Pro and GPT 3.5 Turbo rarely output these long reasoning chains compared to GPT 4 Turbo.

%% file: charts/mmlu_acc_ratio.tex
\begin{figure}\TopFloatBoxes

\begin{floatrow}

\ffigbox[\FBwidth]{
{\caption{Overall accuracy on MMLU with 5-shot prompts and chain-of-thought prompts}\label{fig:mmlu_overall_accuracy}}
{\includegraphics[trim=0 88 0 0, clip, width=0.58\textwidth]{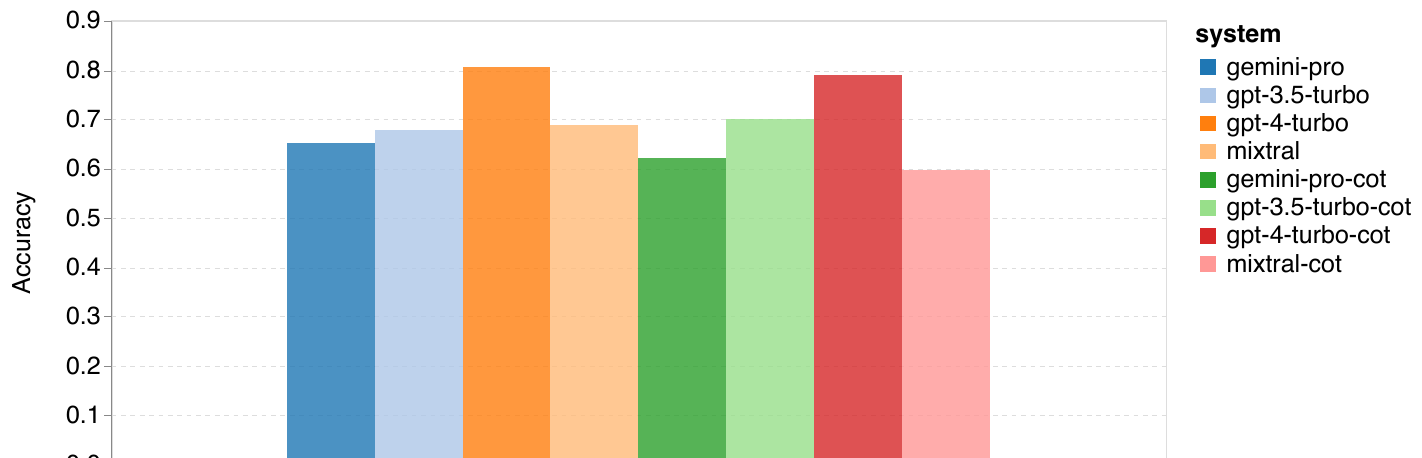}}
}

    \ffigbox[\FBwidth]{
 {\includegraphics[width=0.335\textwidth]{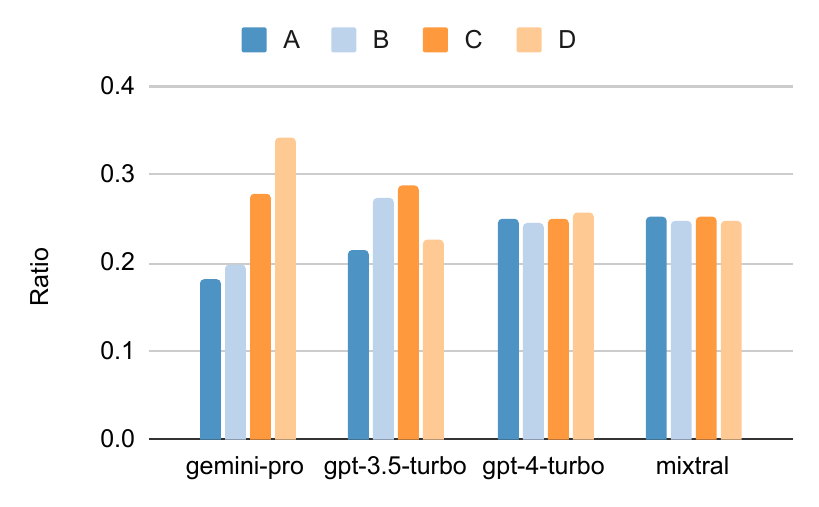}}
    {\caption{Ratio of multiple-choice answers being predicted by models}\label{fig:mmlu_multiple_choice_answer_ratio}}
}

\end{floatrow}
\vspace{-5mm}
\end{figure}

%% file: sections/04_General_Reasoning.tex
\resultsection{General-purpose Reasoning}{https://hub.zenoml.com/report/2575/Gemini\%20BBH}
\label{sec:bbh}

In this category, we focus on 27 diverse reasoning tasks from BIG-Bench Hard \citep{suzgun2022challenging} which consists of arithmetic, symbolic and multilingual reasoning and factual knowledge understanding tasks.
Most of the tasks consist of 250 question-answer pairs, with a few having somewhat fewer.

\subsection{Experimental Details}

\paragraph{Generation Parameters}
We follow standard 3-shot prompts from the Eleuther harness across all models where each question is followed by a chain of thought resulting in a final concluding sentence of ``So the answer is \_\_\_.''.
For hyperparameters, we perform greedy decoding, generating with temperature of 0.

\paragraph{Evaluation}
The Eleuther evaluation harness implementation of BIG-Bench Hard matches the sentence ``So the answer is \_\_\_.'' and extracts the text.
However, we found that for some models, they did not produce this sentence verbatim, even in cases when they generated the correct answer, particularly multiple-choice tasks where the answer is an option chosen from the question text (e.g., ``answer: (B)'').
To remedy this, we modified the matching rule, instead taking the last word of the generated text as the answer of the question only for multiple-choice tasks.

\subsection{Results and Analysis}

For the reasoning tasks, we report the overall performance, performance by question complexity, performance by answer types and performance by BIG-Bench sub-task.

\input{charts/bbh_acc_qlen}

First, we illustrate the overall accuracy in \autoref{fig:reasoning_overall_accuracy}, we can see that Gemini Pro achieves an accuracy slightly lower than that of GPT 3.5 Turbo, and much lower than that of GPT 4 Turbo. In contrast, the Mixtral model achieves somewhat lower accuracy than Gemini Pro.


Based on this overall result, let us dig a little bit deeper into why Gemini might be underperforming. First, we examined \emph{accuracy by the length of the question}, as detailed in \autoref{fig:reasoning_by_length}. We found that Gemini Pro underperformed on longer, more complex questions while the GPT models were more robust to this. This was particularly the case for GPT 4 Turbo, which showed very little degradation even on longer questions, indicating an impressively robust ability to understand longer and more complex queries. GPT 3.5 Turbo fell in the middle with respect to this robustness. Mixtral had low accuracy overall compared to the Gemini GPT models but was notably stable with respect to question length until length 70.

\begin{figure}
\begin{subfigure}[t]{0.47\textwidth}
    \includegraphics[trim=0 20 0 0, clip, width=\textwidth]{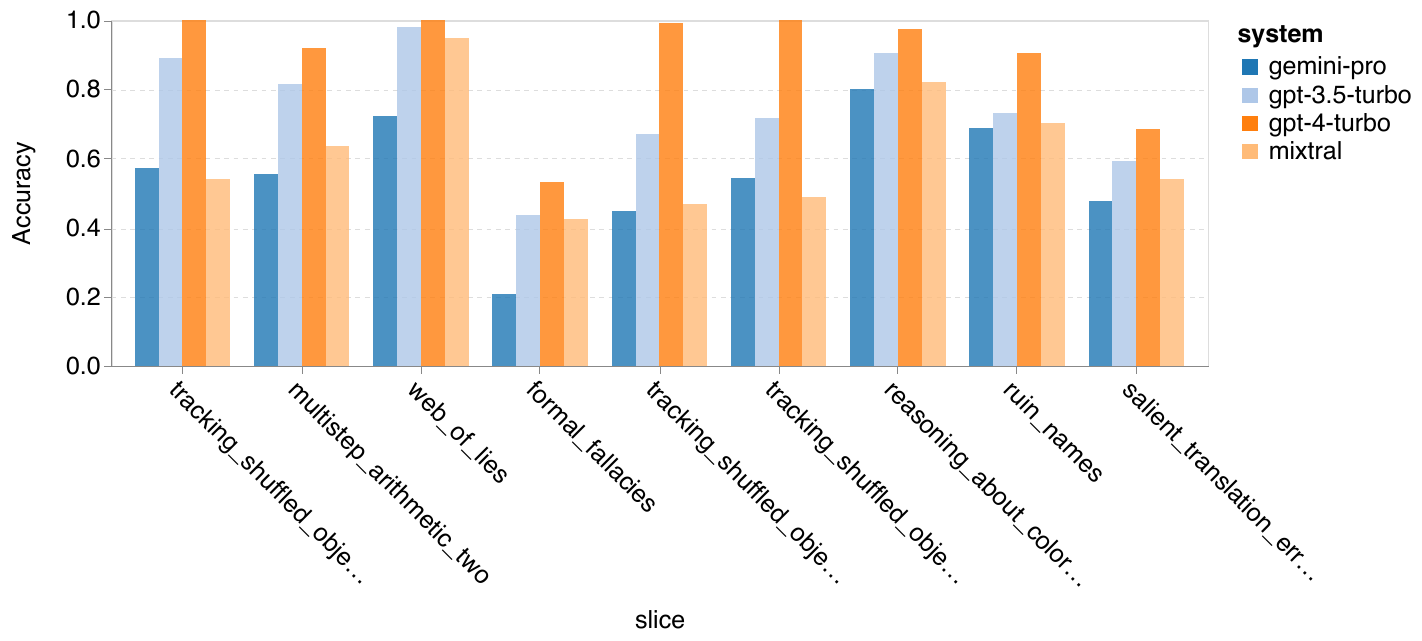}
    \caption{Tasks where GPT 3.5 excels over Gemini Pro \label{fig:gpt3.5_better_gemini}}
\end{subfigure}\hspace{\fill} 
\begin{subfigure}[t]{0.47\textwidth}
    \includegraphics[trim=0 20 0 0, clip, width=\linewidth]{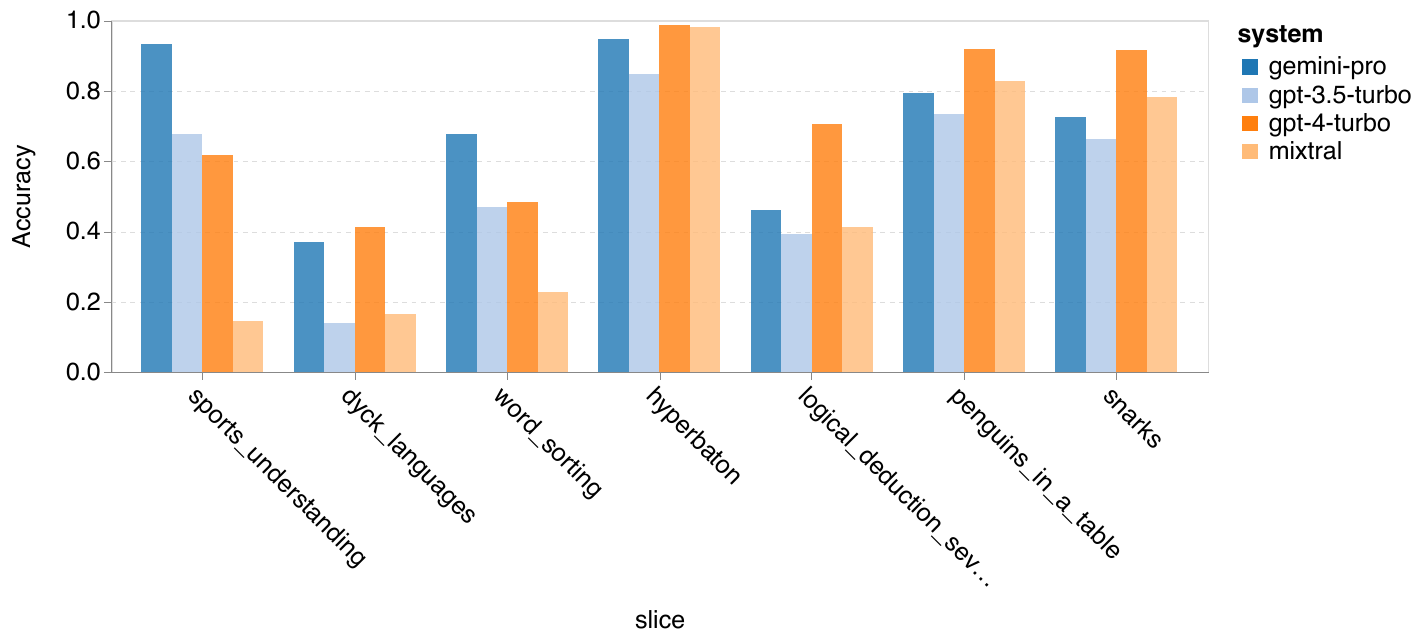}
    \caption{Tasks where Gemini Pro excels over GPT 3.5 \label{fig:gemini_better_gpt3.5}}
\end{subfigure}

\caption{Tasks where Gemini Pro and GPT 3.5 prevail on BBH}
\label{fig:bbh_win_task}
\end{figure}

Next we look at whether there are variations in accuracy by the specific task in BIG-Bench Hard. In \autoref{fig:gpt3.5_better_gemini}, we list the tasks where GPT 3.5 Turbo outperformed Gemini Pro by the largest amount.

We can notice that Gemini Pro is particularly bad at the \texttt{`tracking\_shuffled\_objects'} tasks. These tasks involve keeping track of who has certain objects as they are traded among people, and Gemini Pro often has difficulty keeping the order straight (as  \autoref{tab:ordering}).

\input{tables/tracking_objects}

However, there were a few tasks where Gemini Pro outperformed GPT 3.5 Turbo.
The \autoref{fig:gemini_better_gpt3.5} shows the seven tasks where Gemini Pro outperformed GPT 3.5 Turbo by the largest amount.
These were heterogeneous and included those that required world knowledge (\texttt{sports\_understanding}), manipulating stacks of symbols (\texttt{dyck\_languages}), sorting words in alphabetical order (\texttt{word\_sorting}), detecting sarcasm (\texttt{snarks}) and parsing tables (\texttt{penguins\_in\_a\_table}), among others.

\begin{figure}[ht]
    \centering
    \includegraphics[trim=0 30 0 0, clip, width=0.9\textwidth]{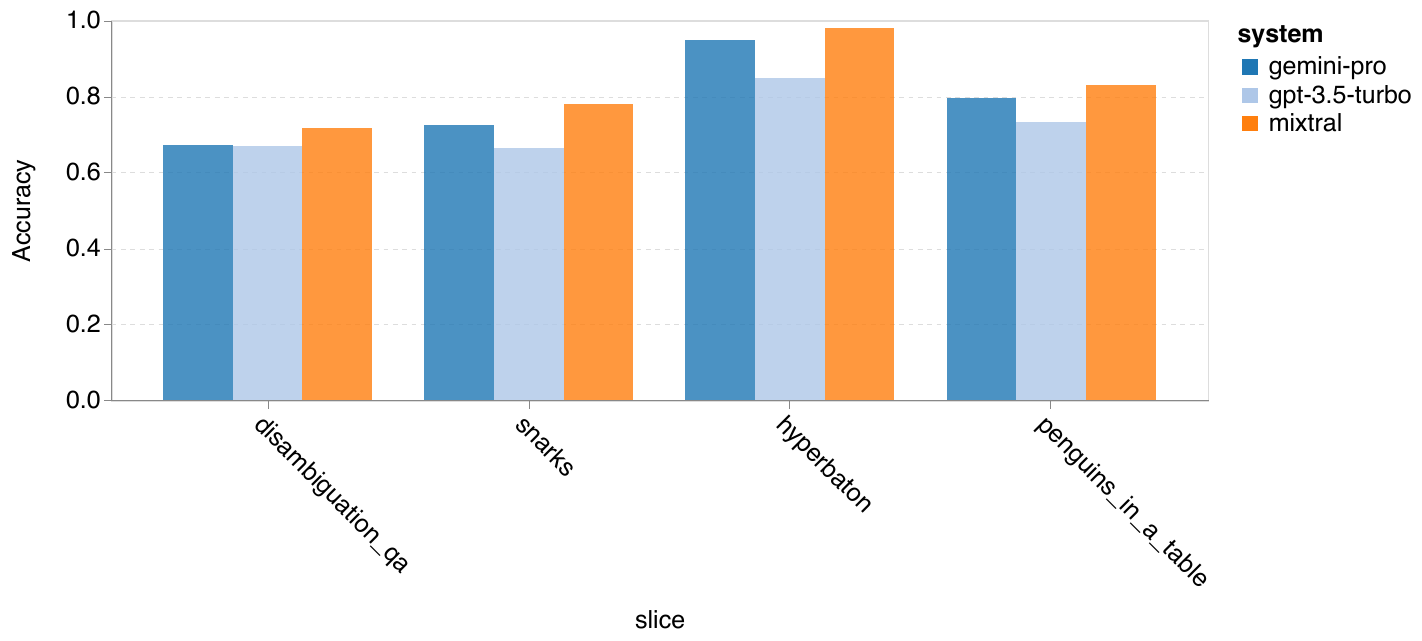}
    \caption{Tasks where Mixtral excels over GPT 3.5 Turbo and Gemini \label{fig:mixtral_better_all}}
\end{figure}

Tasks that mostly require natural language understanding (i.e., semantic understanding, name disambiguation, entity resolution, grammar rules, or sarcasm/humor detection), namely \texttt{salient\_translation\_error\_detection}, \texttt{snarks}, \texttt{hyperbaton}, \texttt{disambiguition\_qa}, and \texttt{ruin\_names}, the Mixtral model performed particularly well, often outperforming both Gemini and GPT 3.5 Turbo (as in \autoref{fig:mixtral_better_all}).


We further investigate the robustness of LLMs across different answer types in the Figure below. We can see that Gemini Pro shows the worst performance in \texttt{Valid/Invalid} answer type which falls under the task \texttt{formal\_fallacies} representing logical deduction from a given context. However, Gemini outperformed all GPT models as well as Mixtral on \texttt{Other} answer types (consisting of the \texttt{word\_sorting} and \texttt{dyck\_language} tasks) which follows a similar line of findings as above i.e., Gemini is particularly good at word rearrangement and producing symbols in the correct order. Also for \texttt{MCQ} and \texttt{Digit} answers, GPT models excel in this genre while Gemini and Mixtral struggles to compete with them.

\begin{wrapfigure}{r}{0.5\textwidth}
    \centering
    \includegraphics[trim=0 20 0 0, clip, width=\textwidth]{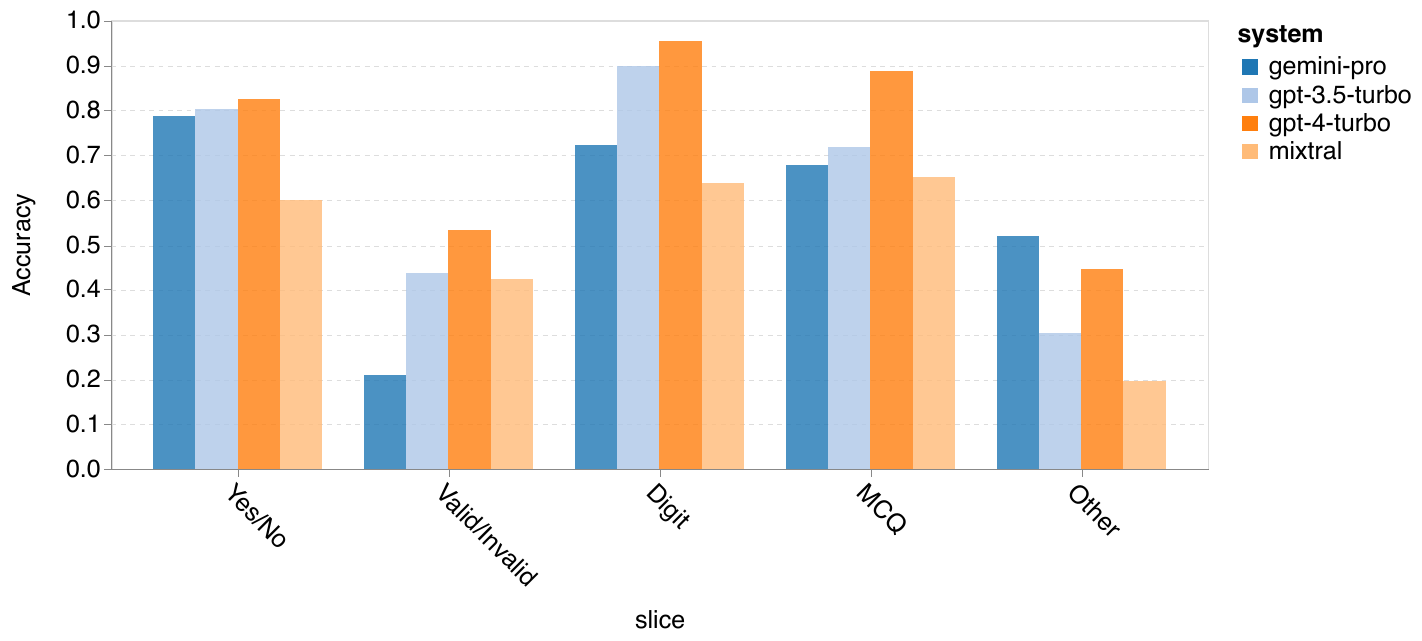}
    \caption{Accuracy by answer types\label{fig:acc_by_answer_type}}
\end{wrapfigure}

In sum, there did not seem to be a particularly strong trend in which tasks one model performed better than the other, so when performing general-purpose reasoning tasks it may be worth trying both the Gemini and GPT models before making a decision on which to use. On the other hand, between Gemini and Mixtral, Mixtral is more reliable in multi-variable reasoning and natural language understanding tasks, among other.

%% file: charts/bbh_acc_qlen.tex
\begin{figure}\TopFloatBoxes

\begin{floatrow}

\ffigbox[\FBwidth]{
{\caption{Overall accuracy on BIG-Bench Hard}\label{fig:reasoning_overall_accuracy}}
{\includegraphics[trim=0 80 0 0, clip, width=0.48\textwidth]{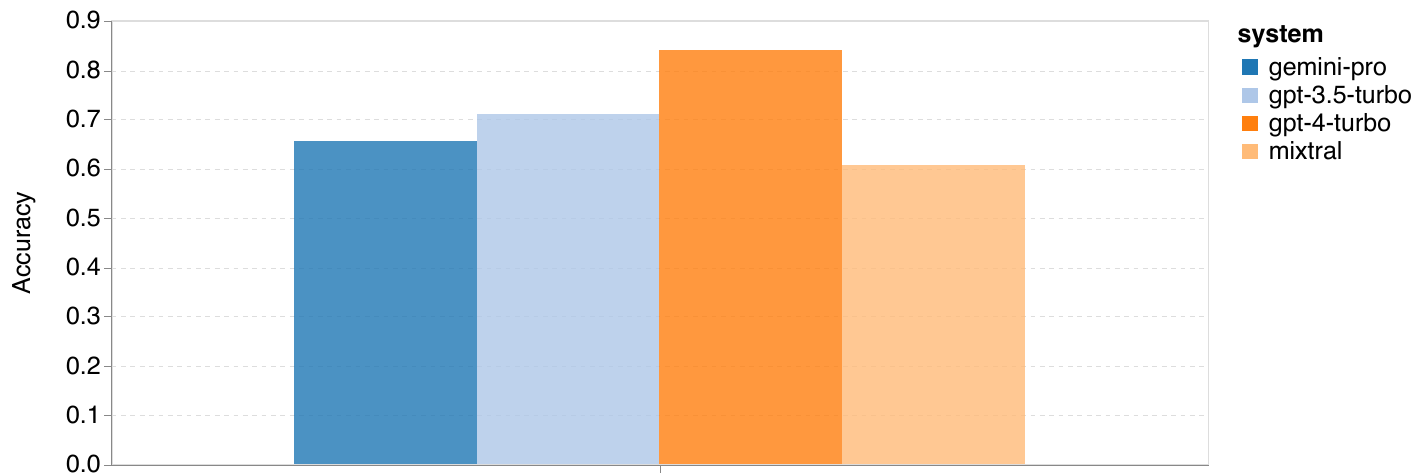}}
}

    \ffigbox[\FBwidth]{
 {\includegraphics[trim=0 20 0 0, clip, width=0.48\textwidth]{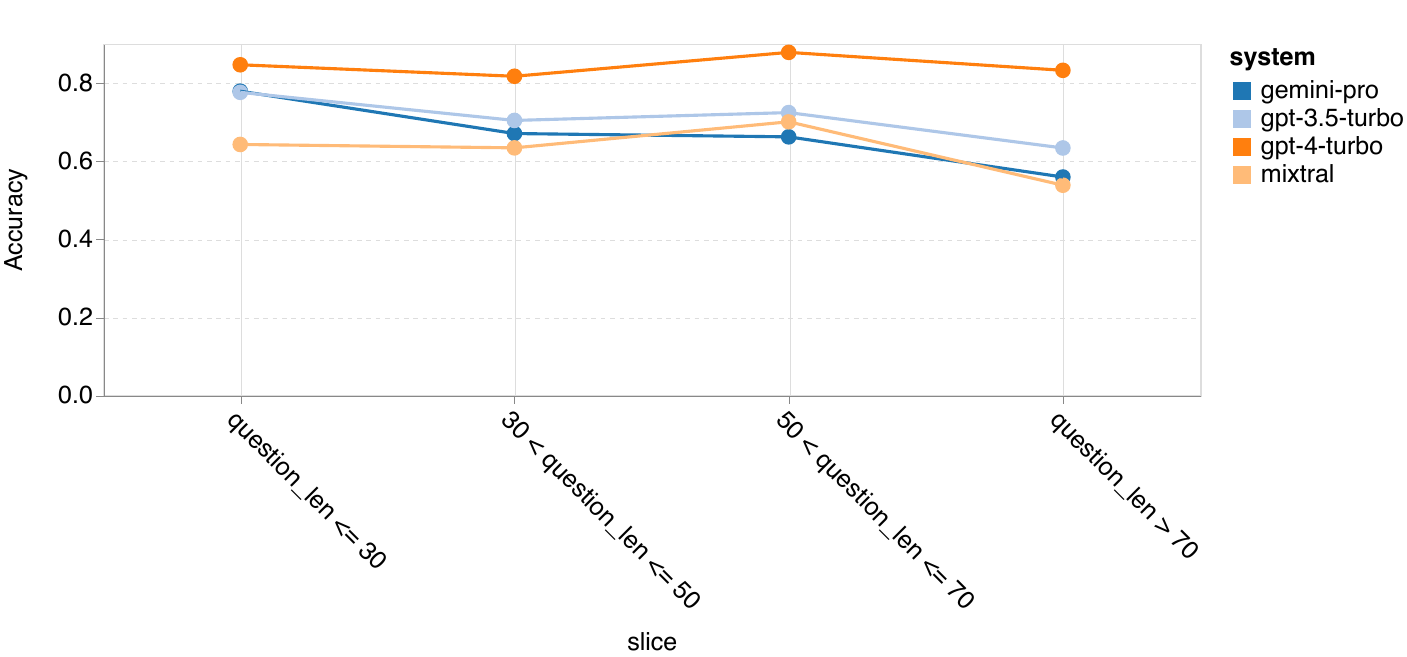}}
    {\caption{Accuracy by question length on BIG-Bench Hard} \label{fig:reasoning_by_length}}
}

\end{floatrow}
\end{figure}

%% file: tables/tracking_objects.tex
\begin{table}[h]
\centering
\caption{Example of Gemini Pro's inability to maintain state in \texttt{tracking\_shuffled\_objects}.\label{tab:ordering}}
\small
\begin{tabular}{p{2cm}|p{11cm}}
\toprule
\textbf{Prompt} & Alice, Bob, and Claire are friends and avid readers who occasionally trade books. At the start of the semester, they each buy one new book: Alice gets Ulysses, Bob gets Frankenstein, and Claire gets Lolita.\newline
As the semester proceeds, they start trading around the new books. First, Claire and Bob swap books. Then, Bob and Alice swap books. Finally, Claire and Bob swap books. At the end of the semester, Bob has\newline
Options:\newline
(A) Ulysses\newline
(B) Frankenstein\newline
(C) Lolita \\
\hline
\textbf{Label} & (B) \\
\hline
\textbf{Output} & (0) At the start: Alice: Ulysses, Bob: Frankenstein, Claire: Lolita.\newline
(1) Claire and Bob swap books: Alice: Ulysses, Bob: Lolita, Claire: Frankenstein.\newline
(2) Bob and Alice swap books: Alice: Frankenstein, Bob: Ulysses, Claire: Lolita.\newline
(3) Claire and Bob swap books: Alice: Frankenstein, Bob: Lolita, Claire: Ulysses.\newline
At the end of the semester, Bob has Lolita. So the answer is (C). \\
\bottomrule
\end{tabular}
\end{table}

%% file: sections/05_math.tex
\resultsection{Mathematics}{https://hub.zenoml.com/report/2773/Gemini\%20Mathematics}
\label{sec:math}

To evaluate the mathematical reasoning ability of the evaluated models, we explore four math word problems benchmarks (1) the grade-school math benchmark, \texttt{GSM8K} \citep{cobbe2021training}, (2) the \texttt{SVAMP} dataset \citep{patel-etal-2021-nlp} with questions generated by varying word-order to check the robust reasoning ability, (3) the \texttt{ASDIV} dataset \citep{miao-etal-2020-diverse} with diverse language patterns and problem types and (4) the \texttt{MAWPS} benchmark \citep{koncel-kedziorski-etal-2016-mawps} consisting of arithmetic and algebraic word problems.

\subsection{Experimental Details}

\paragraph{Generation Parameters}
We consider standard 8-shot chain-of-thought prompts \citep{eval-harness, wei2022chain} where each question in few-shot prompting is associated with a chain of thought for generating the corresponding answer. All models use greedy decoding using a temperature of 0. 

\paragraph{Evaluation}
In evaluation, we make a slight modification to the standard evaluation protocol in the Eleuther harness, which consisted of matching the words ``The answer is'' followed by a numerical output.
We found that all evaluated models had a tendency to output the correct answer even when this specific phrase was not present.
To mitigate this, we are simply taking the last number of the generated text as the answer to the question, which resulted in higher accuracy overall.

\subsection{Results and Analysis}

In this section, we compare the accuracy of Gemini Pro to GPT 3.5 Turbo, GPT 4 Turbo, and Mixtral, on the four math word problems tasks, examining overall performance, performance by question complexity, and performance by chain-of-thought depth. 

\begin{figure}[ht!]
\captionsetup[subfigure]{justification=Centering}

\begin{subfigure}[t]{0.45\textwidth}
    \includegraphics[trim=0 80 0 0, clip, width=\textwidth]{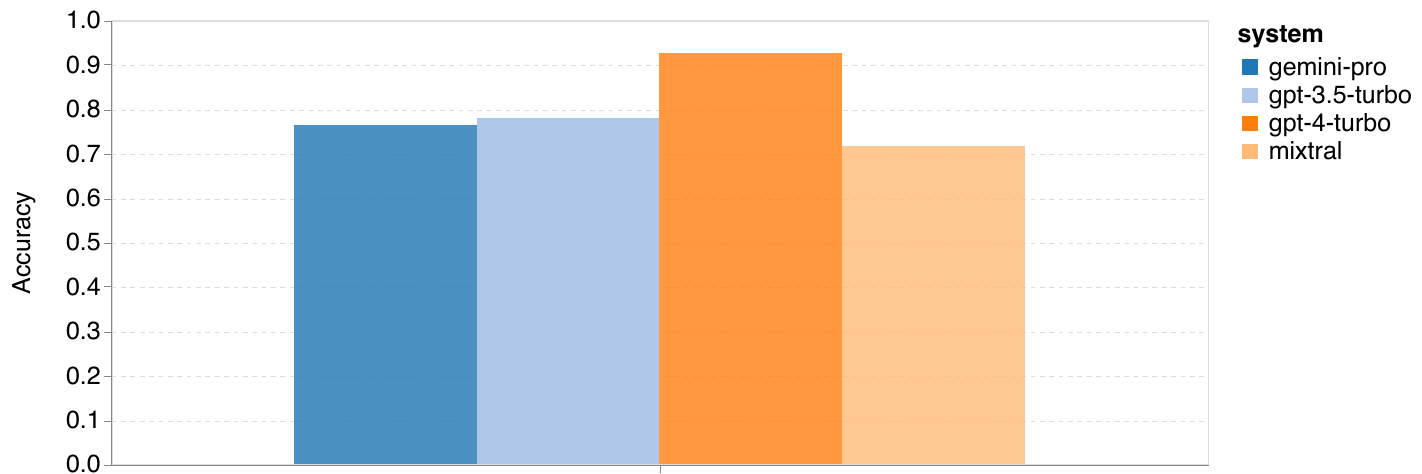}
    \caption{GSM8K}
\end{subfigure}\hspace{\fill} 
\begin{subfigure}[t]{0.45\textwidth}
    \includegraphics[trim=0 80 0 0, clip, width=\linewidth]{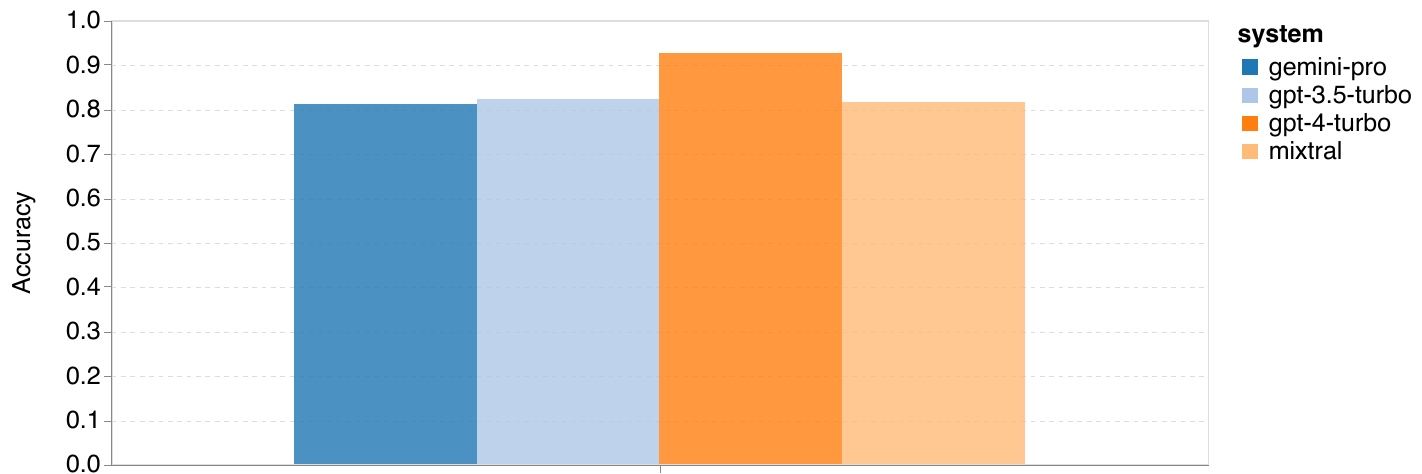}
    \caption{SVAMP}
\end{subfigure}
\bigskip 
\begin{subfigure}[t]{0.45\textwidth}
    \includegraphics[trim=0 80 0 0, clip, width=\linewidth]{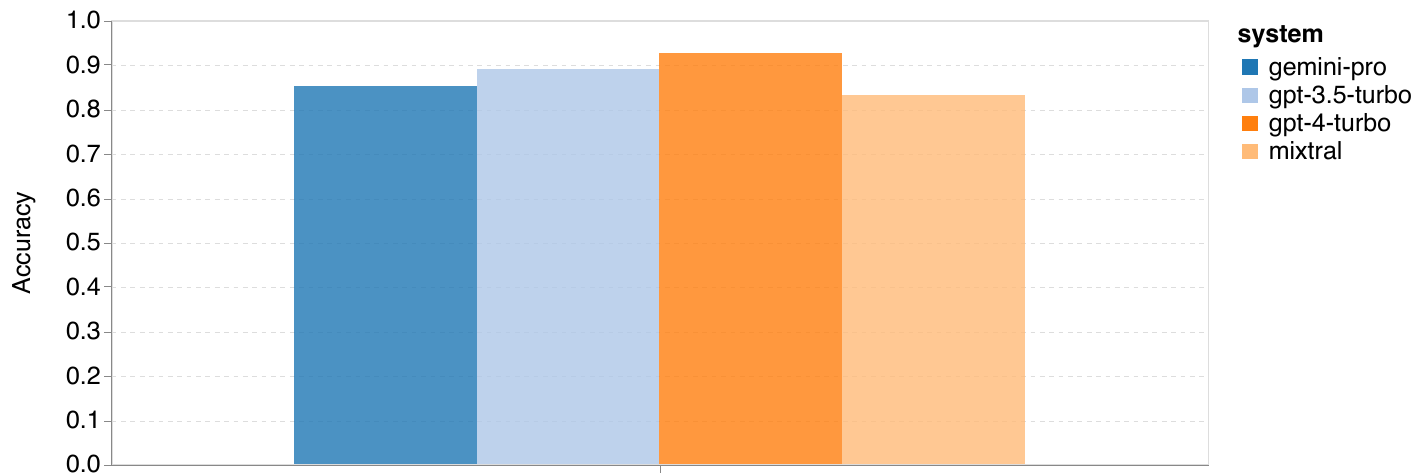}
    \caption{ASDIV}
\end{subfigure}\hspace{\fill} 
\begin{subfigure}[t]{0.45\textwidth}
    \includegraphics[trim=0 80 0 0, clip, width=\linewidth]{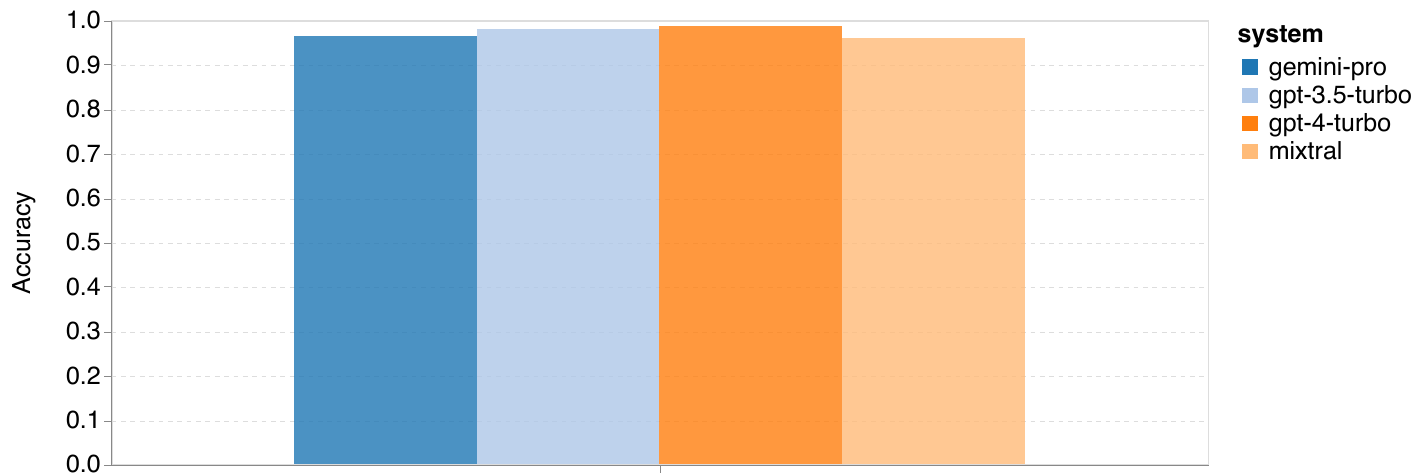}
    \caption{MAWPS}
\end{subfigure}

\caption{Overall accuracy across four mathematical reasoning tasks}
\label{fig:reasoning_accuracy_all_math}
\end{figure}

\begin{figure}[ht!]
\captionsetup[subfigure]{justification=Centering}

\begin{subfigure}[t]{0.45\textwidth}
    \includegraphics[trim=0 20 0 0, clip, width=\textwidth]{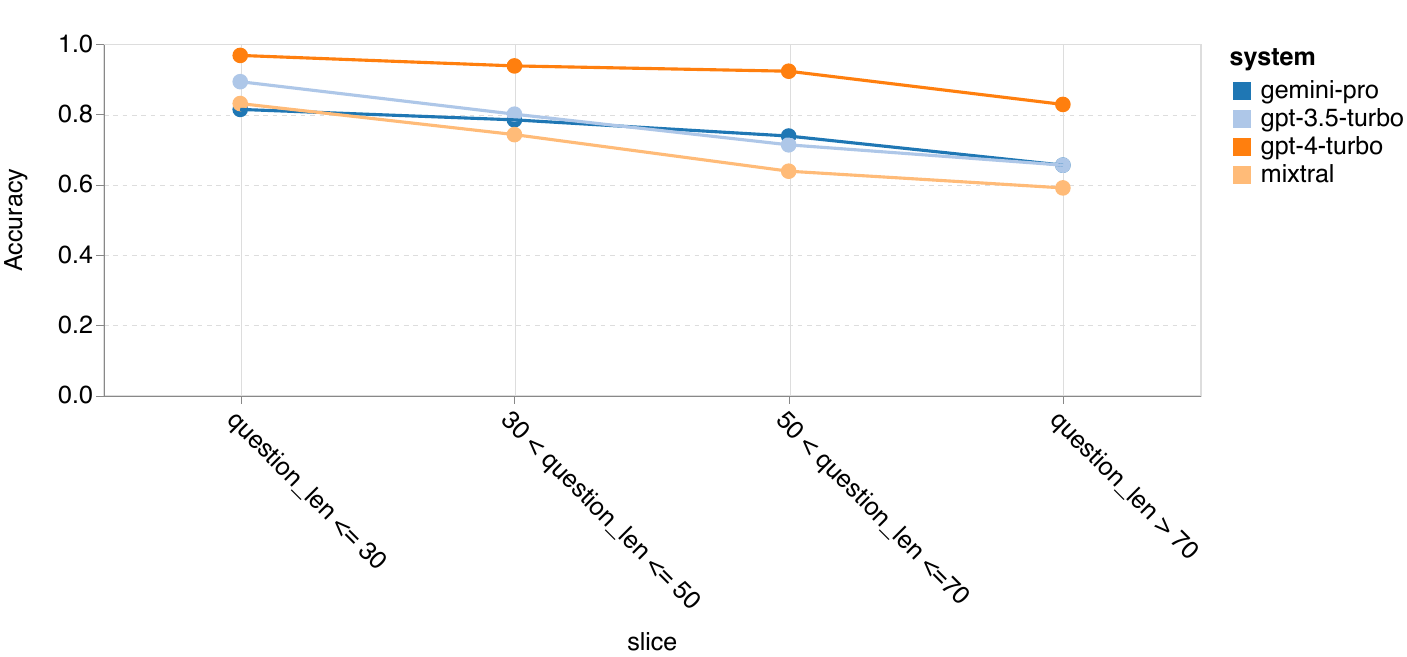}
    \caption{GSM8K}
\end{subfigure}\hspace{\fill} 
\begin{subfigure}[t]{0.45\textwidth}
    \includegraphics[trim=0 20 0 0, clip, width=\linewidth]{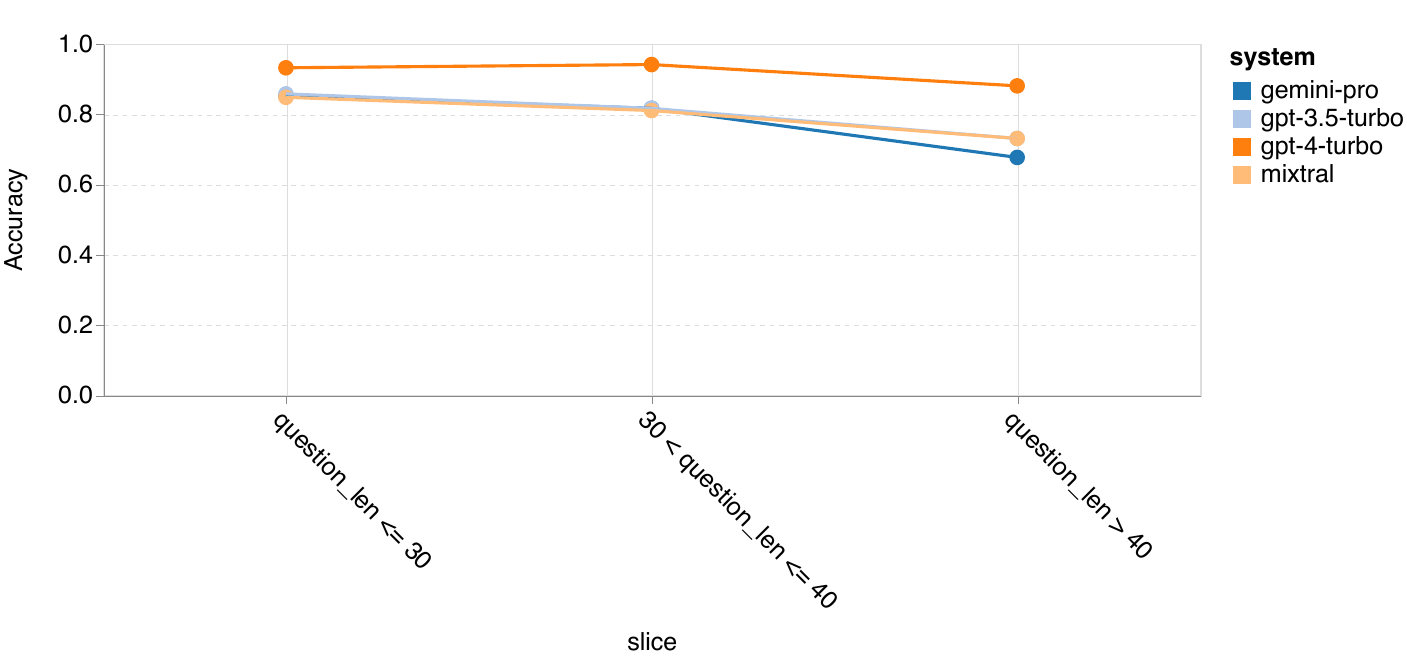}
    \caption{SVAMP}
\end{subfigure}

\bigskip 
\begin{subfigure}[t]{0.45\textwidth}
    \includegraphics[trim=0 20 0 0, clip, width=\linewidth]{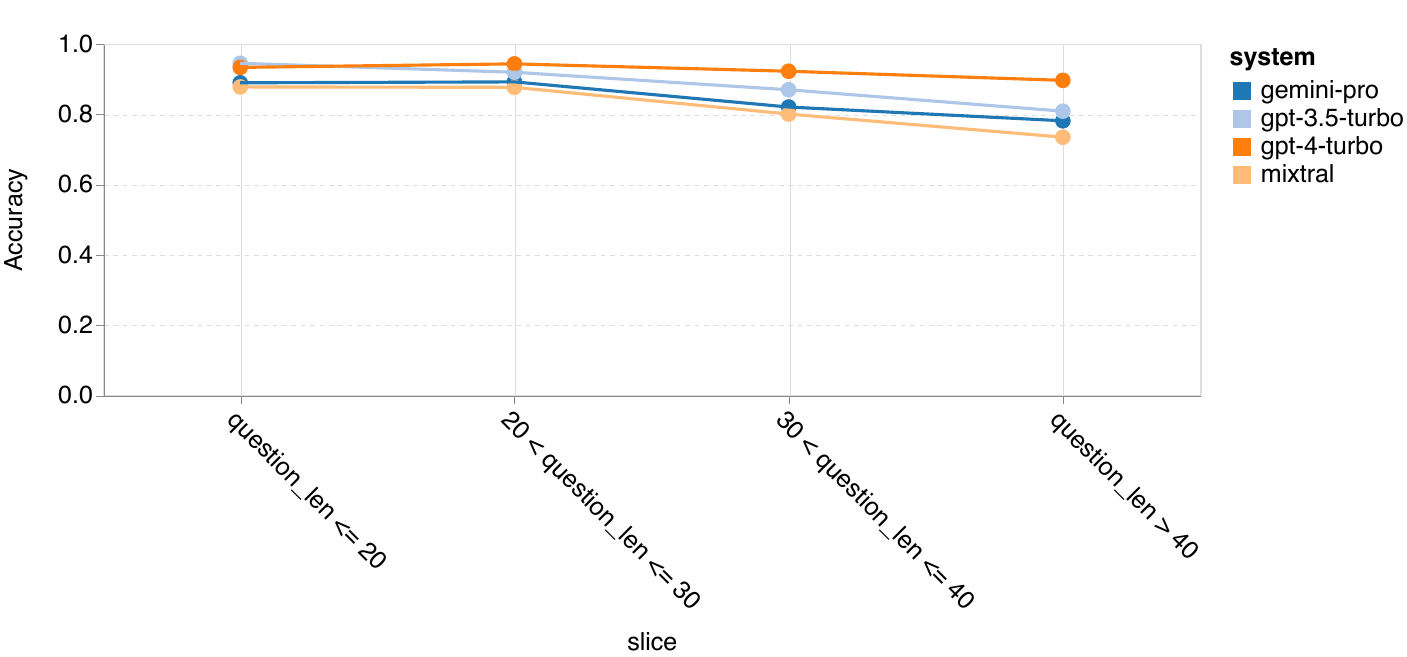}
    \caption{ASDIV}
\end{subfigure}\hspace{\fill} 
\begin{subfigure}[t]{0.45\textwidth}
    \includegraphics[trim=0 20 0 0, clip, width=\linewidth]{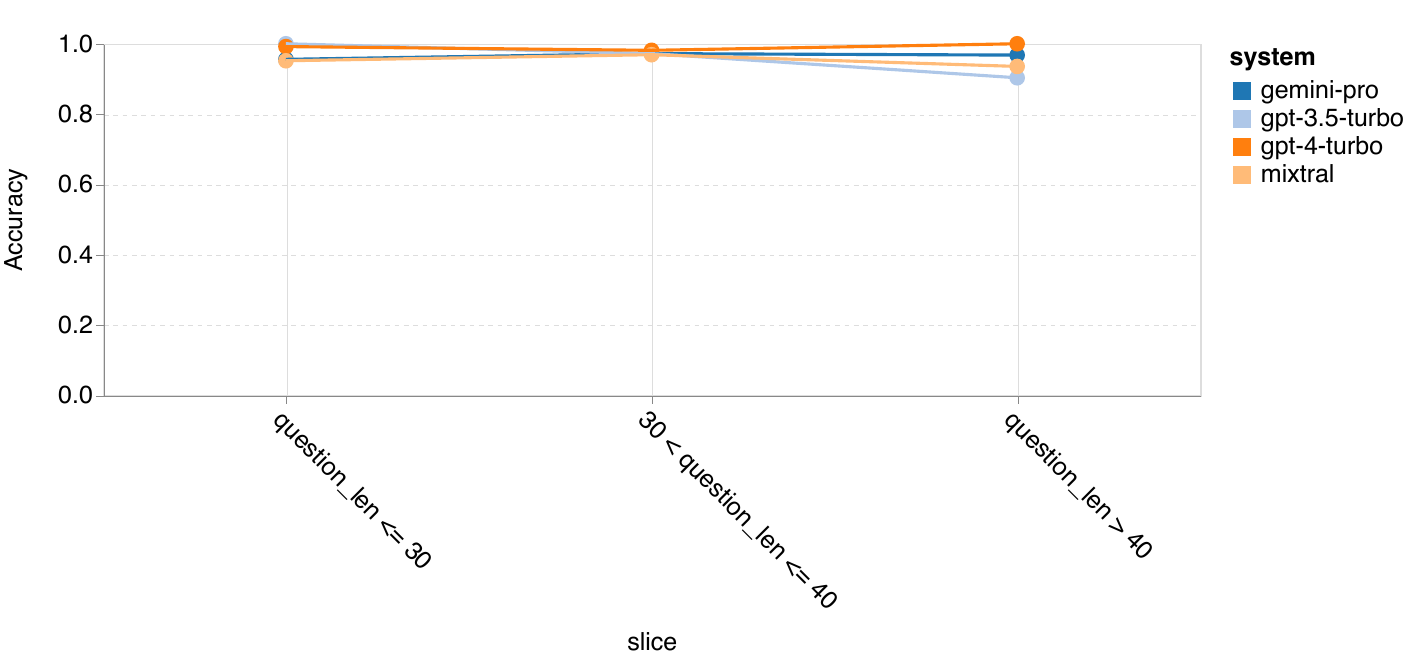}
    \caption{MAWPS}
\end{subfigure}

\caption{Accuracy by question length across four mathematical reasoning tasks}
\label{fig:reasoning_accuracy_by_question_length_all_math}
\end{figure}

First, looking at overall results in the \autoref{fig:reasoning_accuracy_all_math}, we can see that Gemini Pro achieves an accuracy slightly lower than that of GPT 3.5 Turbo, and much lower than that of GPT 4 Turbo on the GSM8K, SVAMP and ASDIV tasks, which all contain diverse language patterns. For the MAWPS task, all models achieve more than 90\% accuracy, although Gemini Pro is still slightly worse than the other models.
In contrast, the Mixtral model achieves slightly lower accuracy compared to Gemini Pro except on the SVAMP task.

Similarly to \autoref{sec:bbh} we break down the results to observe the robustness of each model to question length in \autoref{fig:reasoning_accuracy_by_question_length_all_math}.
As with the reasoning tasks on BIG-Bench Hard, we see a drop-off on longer questions.
As before, GPT 3.5 Turbo outperforms Gemini Pro on shorter questions, but drops off more quickly, with Gemini Pro achieving similar (but still slightly inferior) accuracy on longer questions - except the MAWPS task where Gemini excels over GPT 3.5 on longer questions. Mixtral's degradation is slightly worse than Gemini's except for SVAMP.

\begin{figure}[t!]
\captionsetup[subfigure]{justification=Centering}

\begin{subfigure}[t]{0.45\textwidth}
    \includegraphics[trim=0 20 0 0, clip, width=\textwidth]{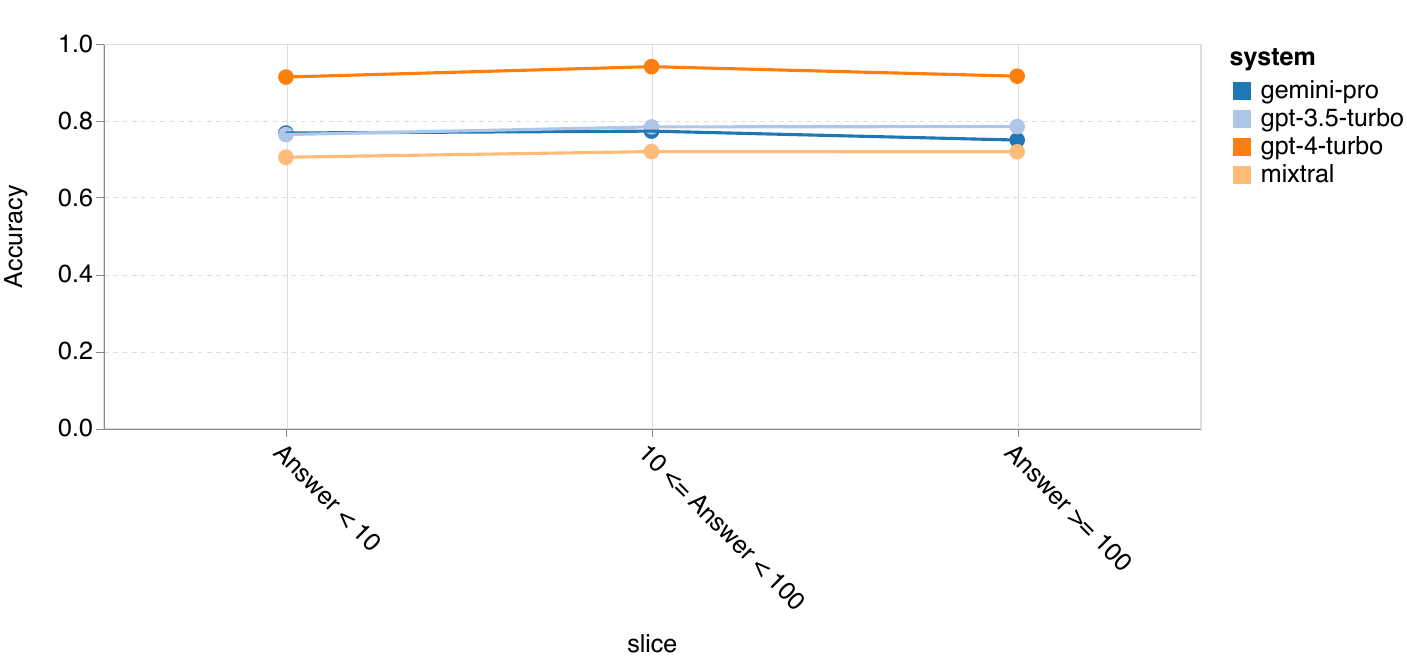}
    \caption{GSM8K}
\end{subfigure}\hspace{\fill} 
\begin{subfigure}[t]{0.45\textwidth}
    \includegraphics[trim=0 20 0 0, clip, width=\linewidth]{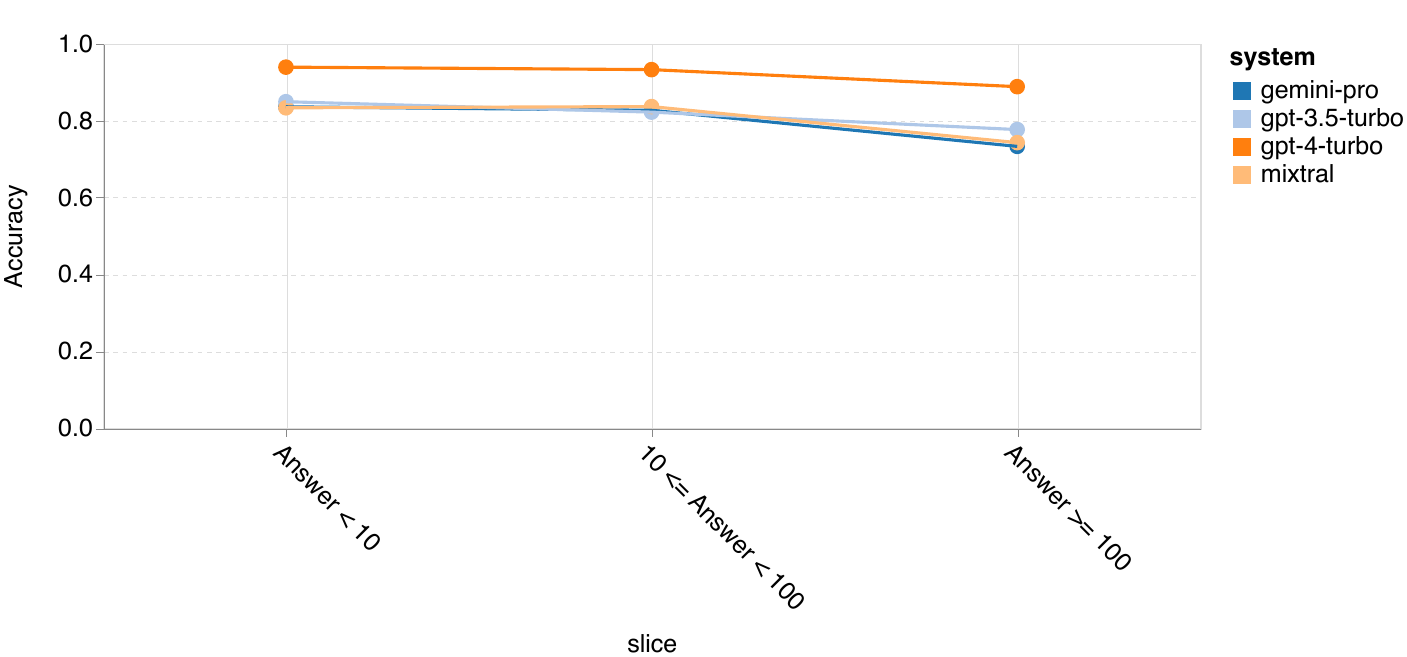}
    \caption{SVAMP}
\end{subfigure}

\begin{subfigure}[t]{0.45\textwidth}
    \includegraphics[trim=0 20 0 0, clip, width=\linewidth]{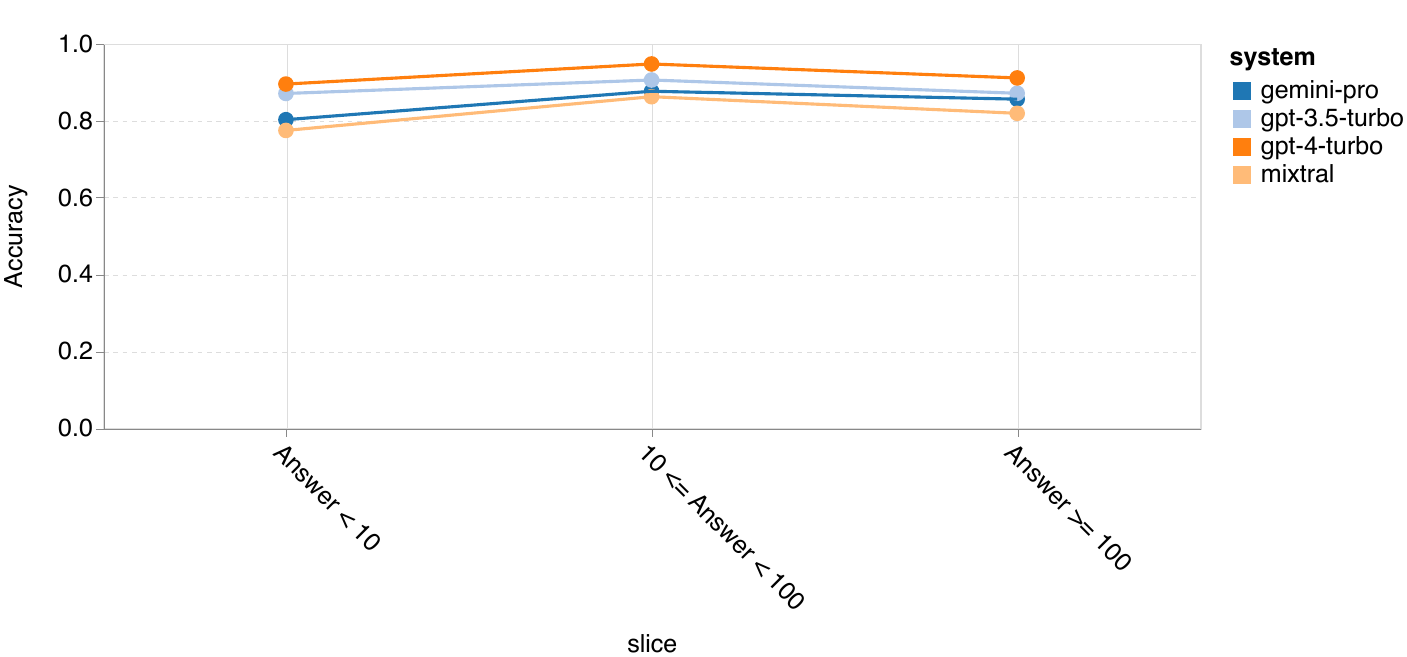}
    \caption{ASDIV}
\end{subfigure}\hspace{\fill} 
\begin{subfigure}[t]{0.45\textwidth}
    \includegraphics[trim=0 20 0 0, clip, width=\linewidth]{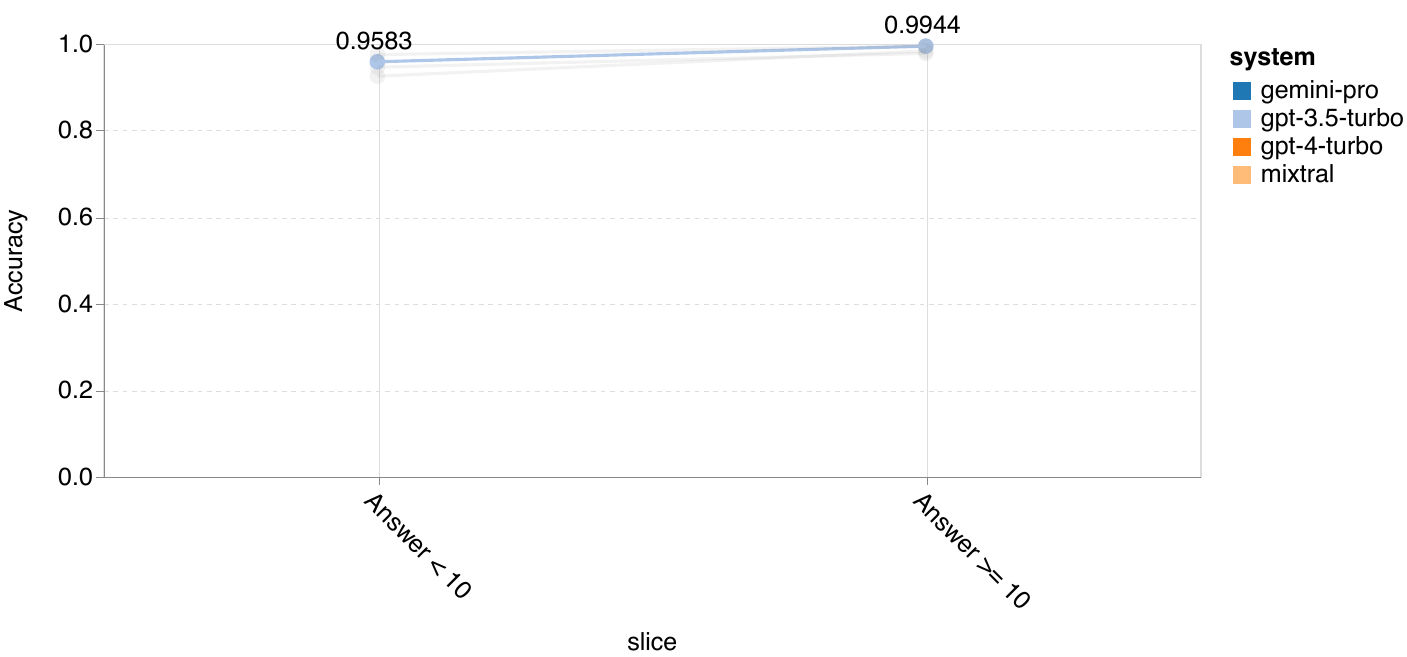}
    \caption{MAWPS}
\end{subfigure}

\caption{Accuracy by number of digits in the answer across four mathematical reasoning tasks}
\label{fig:reasoning_digit_accuracy_all_math}
\end{figure}

\begin{wrapfigure}{r}{0.6\textwidth}
    \vspace{-6mm}
    \centering    
    \includegraphics[trim=0 20 0 0, clip, width=\textwidth]{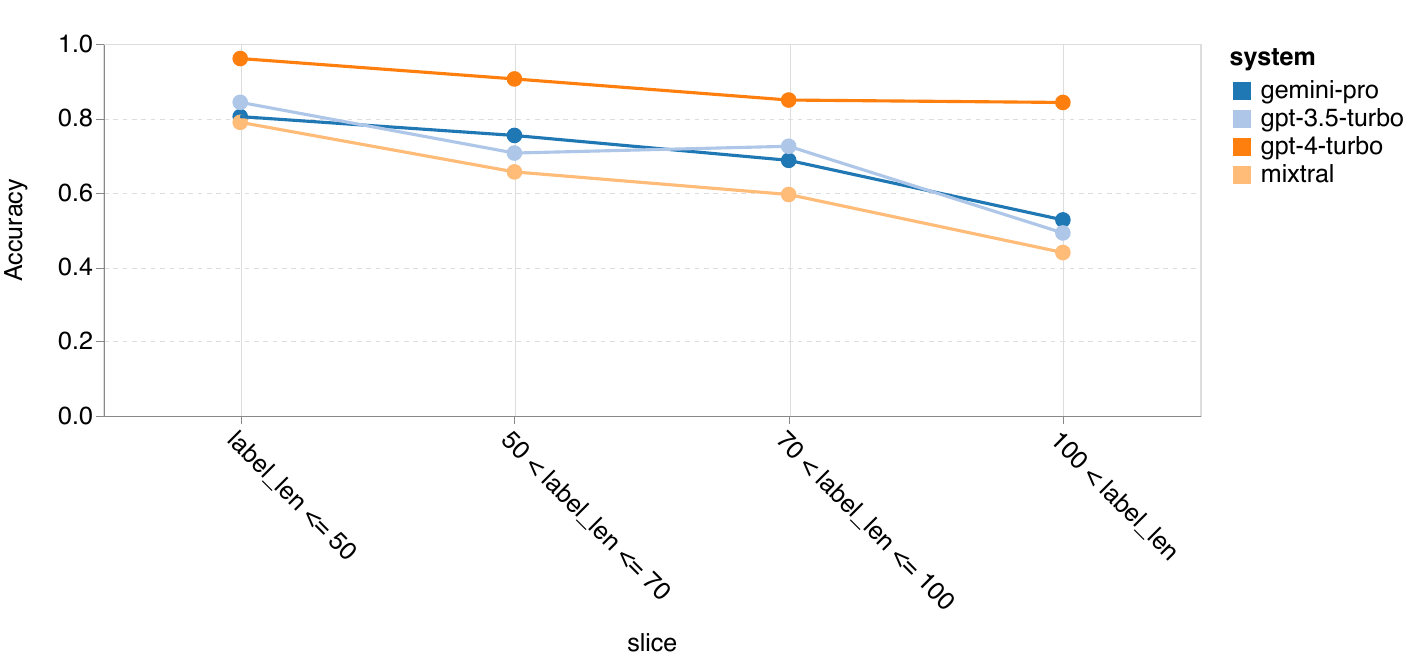}
    \caption{GSM8K accuracy by chain-of-thought length \label{fig:reasoning_accuracy_label_length_gsm}}
    \vspace{-4mm}
\end{wrapfigure}

Additionally, we observe the accuracy of the models when the answer requires longer chains of thought. As shown in \autoref{fig:reasoning_accuracy_label_length_gsm}, GPT 4 Turbo is very robust even when using long reasoning chains, where GPT 3.5 Turbo, Gemini Pro, and Mixtral struggle with increasing COT lengths. In this analysis, we also find that Gemini Pro is superior to GPT 3.5 Turbo in the most complex examples where the COT length is over 100, but underperforms in the shorter examples. In contrast, Mixtral is more affected by longer chain-of-thoughts compared to other models, showing the lowest performance in the most complex examples.

Finally, we investigate the accuracy of the compared models in generating answers with varying numbers of digits. We create three buckets based on the number of digits in the answer, 1, 2, or 3+ (except for the MAWPS task which does not have answers of more than two digits).
As shown in \autoref{fig:reasoning_digit_accuracy_all_math}, GPT 3.5 Turbo appears to be more robust to multi-digit math problems, where Gemini Pro and Mixtral degrades somewhat more on problems with more digits.

To summarize, GPT 4 Turbo is the best model across all math word problem tasks, showing more than 90\% accuracy in all tasks. In contrast, Gemini Pro and Mixtral lag behind all GPT models in this domain, while Gemini shows slightly better reasoning ability than Mixtral.

%% file: sections/06_Code.tex
\resultsection{Code Generation}{https://hub.zenoml.com/report/2641/Gemini\%20Code}
\label{sec:code} 

In this category, we examine the models' coding abilities using two code generation datasets HumanEval \citep{chen2021evaluating} and ODEX \citep{wang2022execution}.
The former tests basic code understanding on a limited set of functions from the Python standard library, while the latter tests the ability to use a broader set of libraries from the entire Python ecosystem.
Both of them take as input a human-written task description in English (often with test cases).
These problems evaluate comprehension of language, algorithmic understanding, and elementary mathematics. 
Overall, HumanEval has 164 test samples, and ODEX has 439 test samples.

\subsection{Experimental Details}

\paragraph{Generation Parameters}
We follow the standard zero-shot code evaluation pipeline provided by the ODEX\footnote{\url{https://github.com/zorazrw/odex}}.
We generate with temperature 0.0, which demonstrated the best performance for all models, compared to other temperatures.
We use a prompt of  ``Complete the following python3 function: '' to ensure that the models' output fits the desired format.

\paragraph{Evaluation}
We perform execution-based evaluation, measuring the Pass@1 metric, which determines whether a single model output passes test cases \citep{chen2021evaluating}.
Since code generation is evaluated in a zero-shot fashion, the model may inevitably output code that does not conform to our input format well.
Therefore, we perform rudimentary post-processing to make the output code fit into the final verification pipeline as much as possible, including the removal of markdown code blocks, the extraction of function implementations and the truncation of natural language sentences following the code.
We do not automatically fix missing library imports.

\subsection{Results and Analysis}

In this section, we examine the overall performance and present a few examples.

\begin{figure}[th]
\begin{subfigure}[t]{0.48\textwidth}
    \includegraphics[trim=0 80 0 0, clip, width=\textwidth]{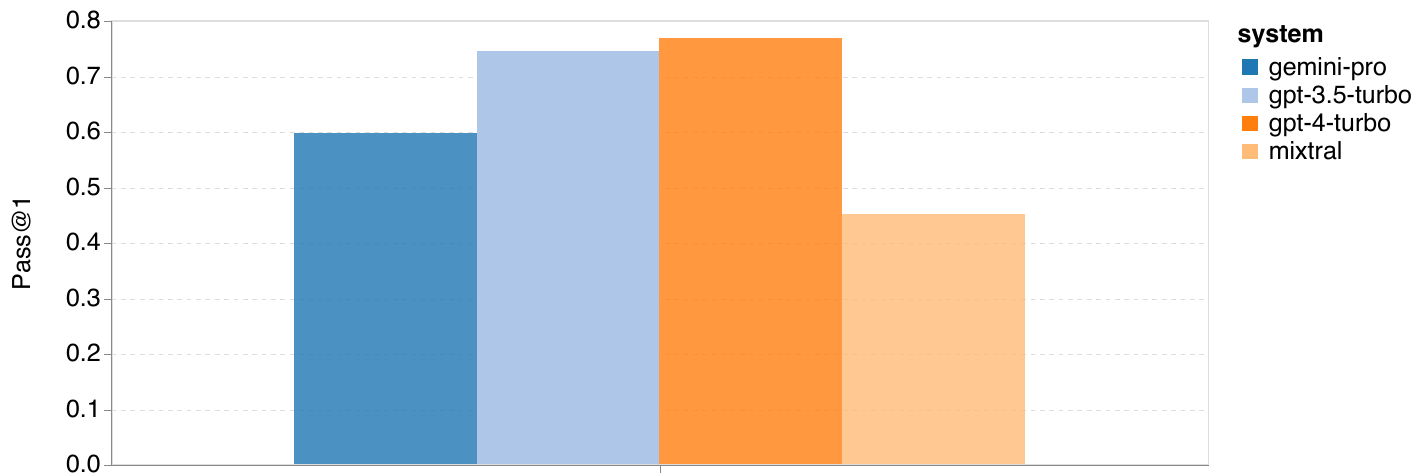}
    \caption{HumanEval}
\end{subfigure}\hspace{\fill} 
\begin{subfigure}[t]{0.48\textwidth}
    \includegraphics[trim=0 80 0 0, clip, width=\linewidth]{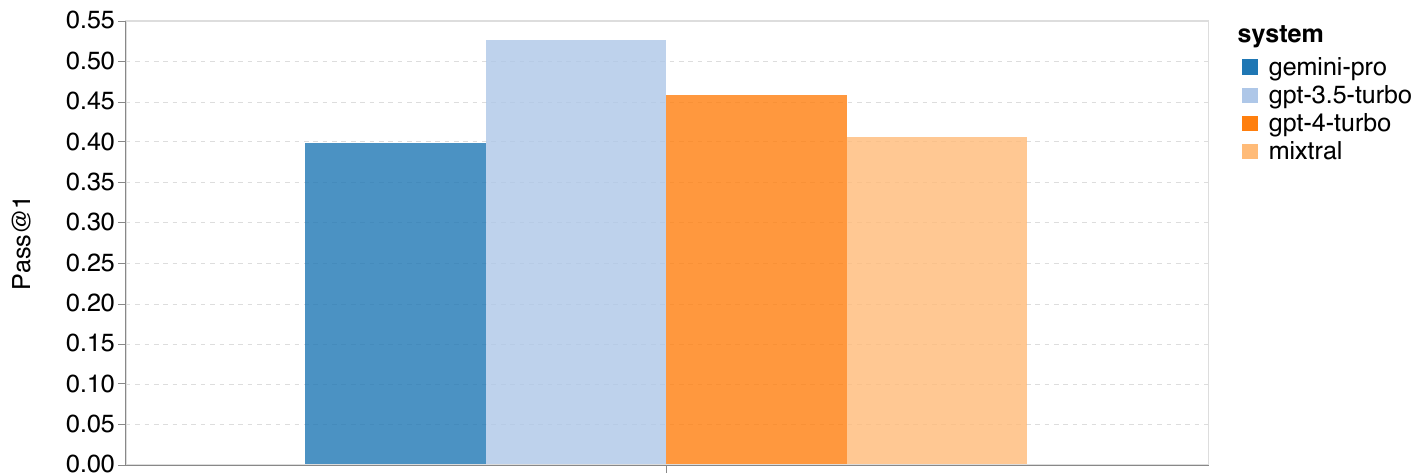}
    \caption{ODEX}
\end{subfigure}
\caption{Overall accuracy on code generation tasks}
\label{fig:code_overall_accuracy}
\end{figure}

First, from the results shown in \autoref{fig:code_overall_accuracy}, we can see that Gemini Pro achieves a Pass@1 lower than GPT 3.5 Turbo and GPT 4 Turbo on both tasks.
On HumanEval, Gemini Pro significantly outperforms Mixtral, and on ODEX Mixtral and Gemini Pro have roughly equivalent accuracy.%
\footnote{Note that the 59.8\% accuracy that Gemini Pro achieved on HumanEval is significantly lower than that reported by \citet{gemini23gemini}. We suspect the difference may be due to differences in prompting techniques, and will continue examining the issue.}

\begin{figure}[t!]
\centering
\begin{subfigure}[t]{0.47\textwidth}
    \includegraphics[trim=0 20 0 0, clip, width=\textwidth]{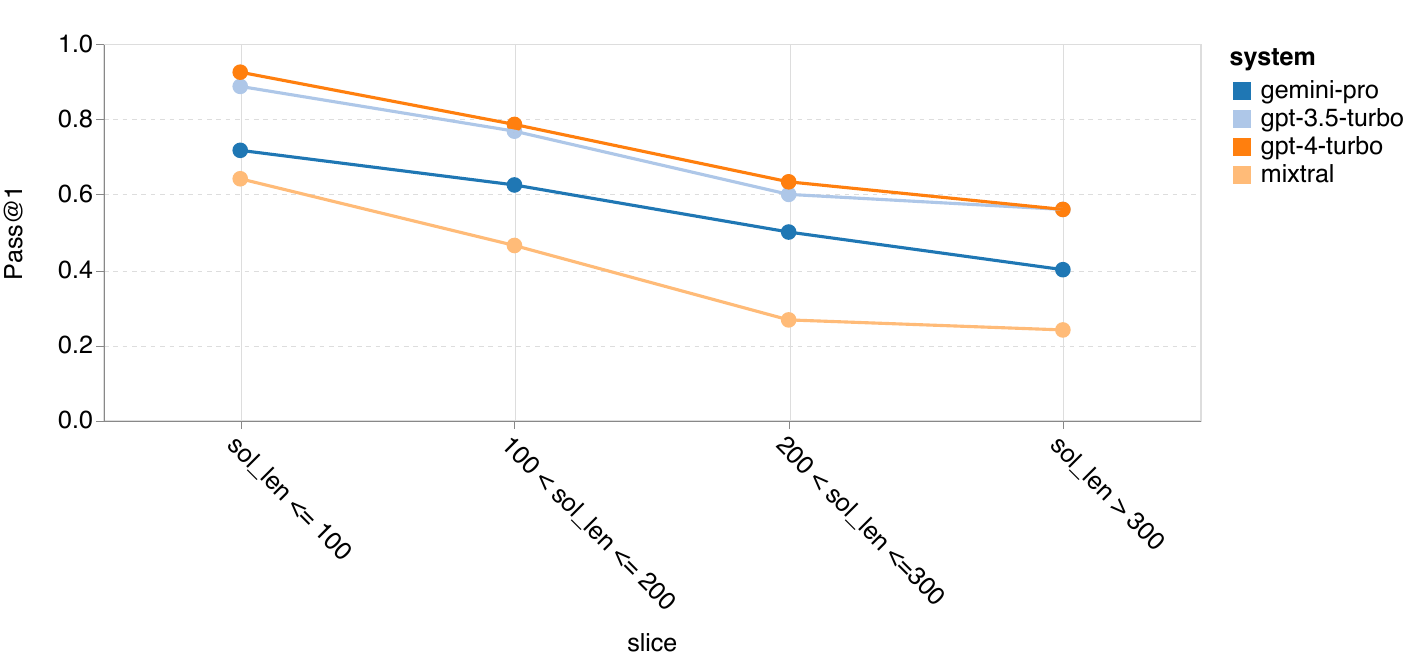}
    \caption{Accuracy by gold solution length on HumanEval}
    \label{fig:code_comp_len}
\end{subfigure}\hspace{\fill} 
\begin{subfigure}[t]{0.47\textwidth}
    \includegraphics[trim=0 20 0 0, clip, width=\linewidth]{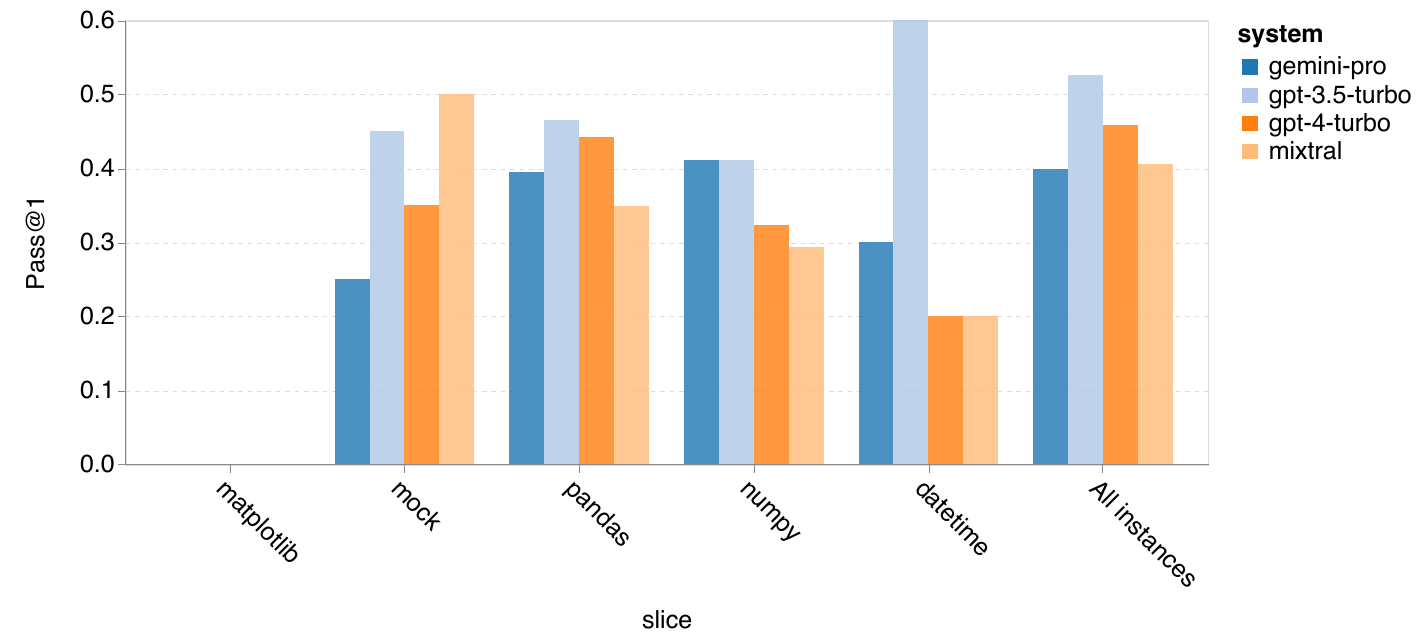}
    \caption{Accuracy by used libraries on ODEX}
    \label{fig:code_comp_lib}
\end{subfigure}
\caption{Comparison of Pass@1 w.r.t. gold solution length and the libraries used by gold solution}
\label{fig:code_comp_len_lib}
\end{figure}

Second, we analyze the relationship between the gold solution length and the model performance in \autoref{fig:code_comp_len}.
The solution length can partly indicate the difficulty of solving the corresponding code generation task. We find that even though Gemini Pro achieves comparable Pass@1 with GPT 3.5 when the solution length is below 100 (e.g., easier cases), it falls behind by large margins when the solution becomes longer.
This is an interesting contrast to the results from previous sections, where we found that in general Gemini Pro performed robustly with respect to longer inputs and outputs on English language tasks.

On the ODEX benchmark all models achieved lower accuracy than HumanEval, with a significant portion of the errors being due to the models failing to import libraries that they were using, or using non-current APIs.
We also present the analysis of how the libraries required in each solution affect the model performance in \autoref{fig:code_comp_lib}. Gemini Pro performs slightly worse than GPT 3.5 on most library-used cases, such as mock, pandas, numpy, and datetime.

Finally, we show several concrete examples of failure cases where Gemini Pro performs worse in code generation than GPT 3.5.
First, we noticed that Gemini is somewhat worse at correctly choosing functions and arguments from the Python API. For instance, given this prompt:

\begin{minted}{python}
def f_3283984():
    """decode a hex string '4a4b4c' to UTF-8."""
\end{minted}

Gemini Pro generated the following code, which results in a type mismatch error:

\begin{minted}{python}
    bytes(bytearray.fromhex('4a4b4c'), 'utf-8')
\end{minted}

In contrast, GPT 3.5 Turbo used the following code, which achieves the desired result:

\begin{minted}{python}
    hex_string = '4a4b4c'
    decoded_string = bytes.fromhex(hex_string).decode('utf-8')
    return decoded_string
\end{minted}

Further, Gemini Pro had a higher proportion of mistakes where the implemented code was syntactically correct but did not correctly match with a more complex intent.
For instance, with respect to the following prompt:

\begin{minted}{python}
from typing import List

def remove_duplicates(numbers: List[int]) -> List[int]:
    """From a list of integers, remove all elements that occur more than once.
    Keep order of elements left the same as in the input.
    >>> remove_duplicates([1, 2, 3, 2, 4])
    [1, 3, 4]
    """
\end{minted}

Gemini Pro created an implementation that just extracts the unique numbers without removing those that appear more than once.

\begin{minted}{python}
    seen_numbers = set()
    unique_numbers = []
    for number in numbers:
        if number not in seen_numbers:
            unique_numbers.append(number)
            seen_numbers.add(number)
    return unique_numbers
\end{minted}

%% file: sections/07_MT.tex
\resultsection{Machine Translation}
{https://hub.zenoml.com/report/2740/Gemini\%3A\%20Flores\%20Translation\%20Evaluation}
\label{sec:mt} 

This set of experiments evaluates the models' multilingual ability, specifically their ability to translate between various language pairs, using the FLORES-200 machine translation benchmark \citep{team2022language}. 
We focus on a diverse subset of 20 languages used by the analysis of \citet{robinson2023chatgpt}, which encompass various levels of resource availability and translation difficulty.
We evaluate on the 1012 sentences from the test set for all the chosen language pairs. As the first step of this study, we limited our scope to translations from English to other languages (ENG$\rightarrow$X) only.%

\subsection{Experimental Details}

\paragraph{Generation Parameters}

We use a five-shot prompting strategy (zero shot results are also noted in \autoref{sec:mt_addn_exps}), specifically the prompts proposed by \citet{gao2023design} designated in \autoref{tab:translation_prompts}.
Our experimental setup employed a top\_p value of 1, a temperature of 0.3, a context\_length of -1, and max\_tokens 500, which we found to generally achieve good performance for translation.

\paragraph{Evaluation}

To evaluate the outputs, we utilized sentence level averaged \textit{chrF2++}, leveraging the implementation provided by sacreBLEU \citep{post2018call}.
This standard metric is based on character and word $n$-gram overlap between the system output and the reference sentence. We compute the sentence level chrF scores.
For simplicity, we refer to this metric as chrF in our discussion \citep{popovic2017chrf++}.

\subsection{Results and Analysis}

\input{tables/mt-5}

\paragraph{Overall Performances} 

In \autoref{table:mt2}, we conduct a comparative analysis of Gemini Pro, GPT 3.5 Turbo, GPT 4 Turbo, and Mixtral.
We also compare against established translation-specific systems like Google Translate\footnote{http://translate.google.com}, and NLLB-MoE \citep{team2022language}, the leading open-source machine translation (MT) model known for its extensive language coverage.

The results indicate that Google Translate generally outperforms other models in languages that it supports, excelling in 10 languages.
It is followed by NLLB, which excels on 6 languages.
Gemini Pro provided impressive accuracy on several languages, even having the best accuracy on 3: South Levantine Arabic, Romanian, and Mesopotamian Arabic (surpassing not only GPT 3.5 Turbo, but also GPT 4 Turbo).

However, on average, the results indicate that the general-purpose language models showed competitive performances but have not yet surpassed the dedicated machine translation systems in translation into non-English languages.

\begin{figure}[ht]
    \centering
    \includegraphics[width=0.8\textwidth]{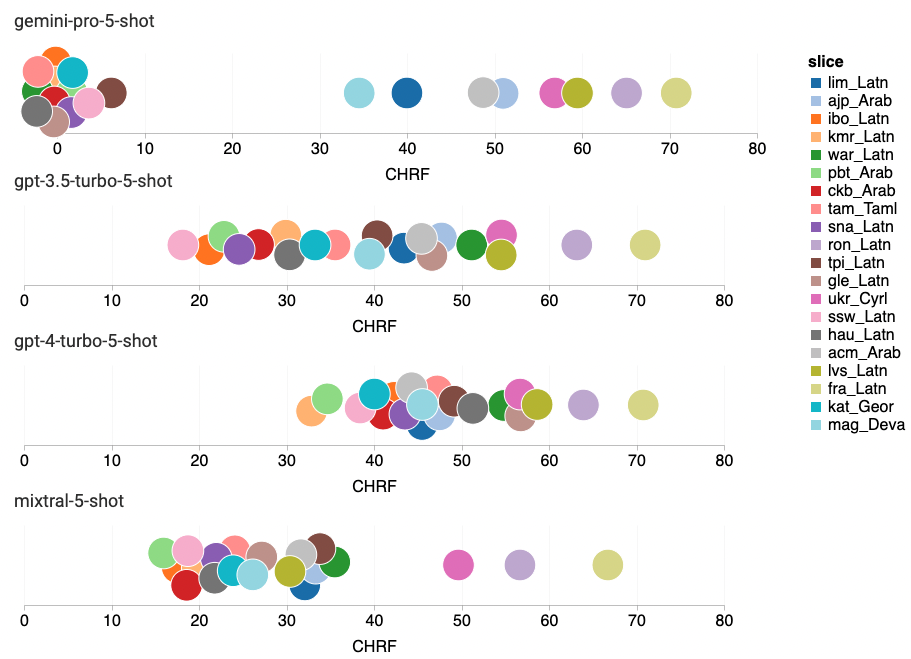}
    \caption{Machine translation performance (chrF (\%) scores) by language pairs for 5-shot prompt.}
    \label{fig:mt_bubble}
\end{figure}

\autoref{fig:mt_bubble} illustrates the comparative performance of language models across language pairs for the 5-shot prompt. 
GPT 4 Turbo showed a consistent deviation of performance with NLLB relative to GPT 3.5 Turbo and Gemini Pro. This reflects the findings in the literature on GPT 4 Turbo's multilingual performance \citep{openai2023gpt4}. GPT 4 Turbo also offered larger improvements for low-resource languages (as measured by \citet{team2022language}), whereas for high-resource languages performance was similar between the LLMs.
In comparison, Gemini Pro outperforms both GPT 3.5 Turbo and GPT 4 Turbo on 5 out of 20 languages, and achieved the top performances on 3 languages.
Mixtral underperformed the other models across language pairs.

\begin{figure}[th]
    \centering
    \includegraphics[trim=0 60 0 0, clip, width=0.8\textwidth]{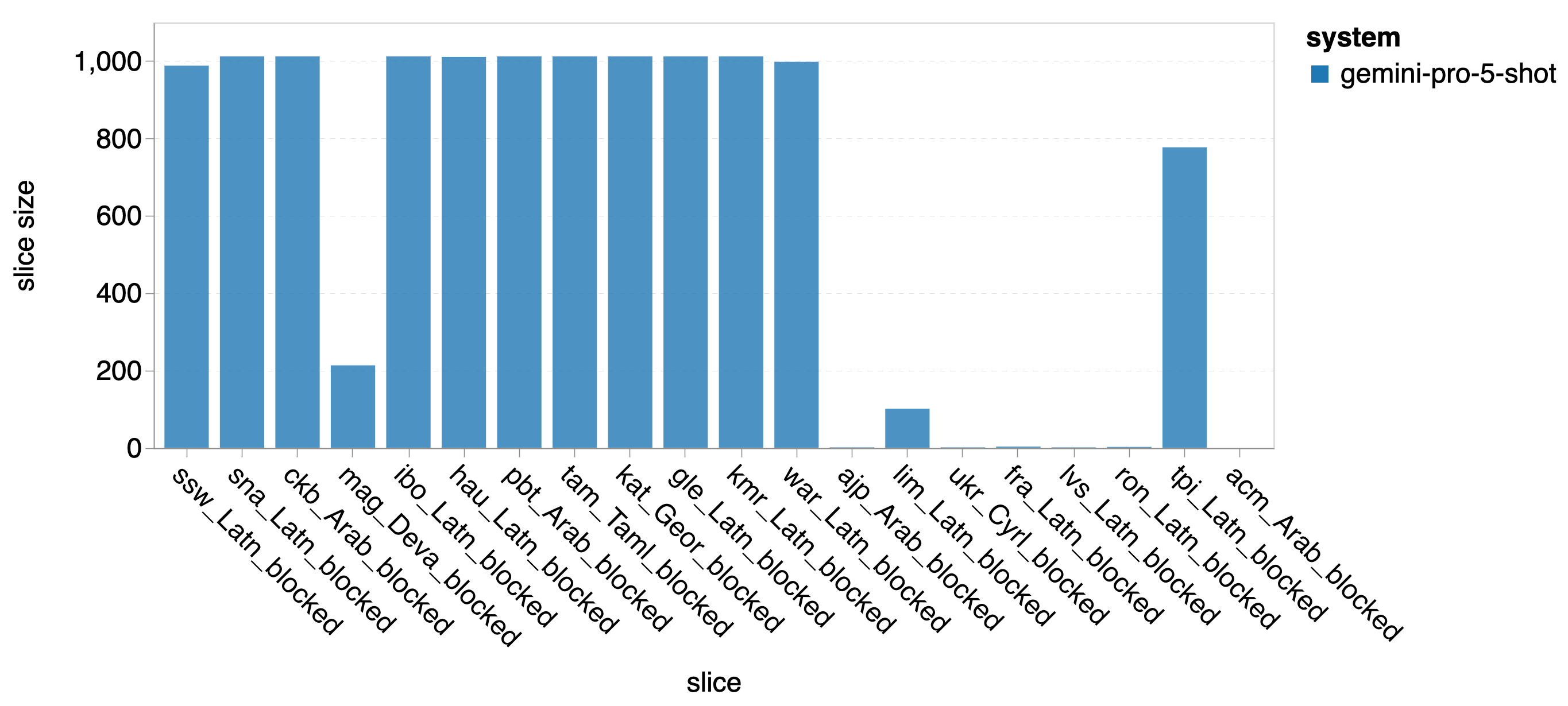}
    \vspace{-3mm}
    \caption{Number of samples that are blocked by Gemini Pro 5-shot}
    \label{fig:mt_blocked_five}
\end{figure}


\begin{figure}[h]
    \centering
    \includegraphics[trim=0 60 0 0, clip, width=0.95\textwidth]{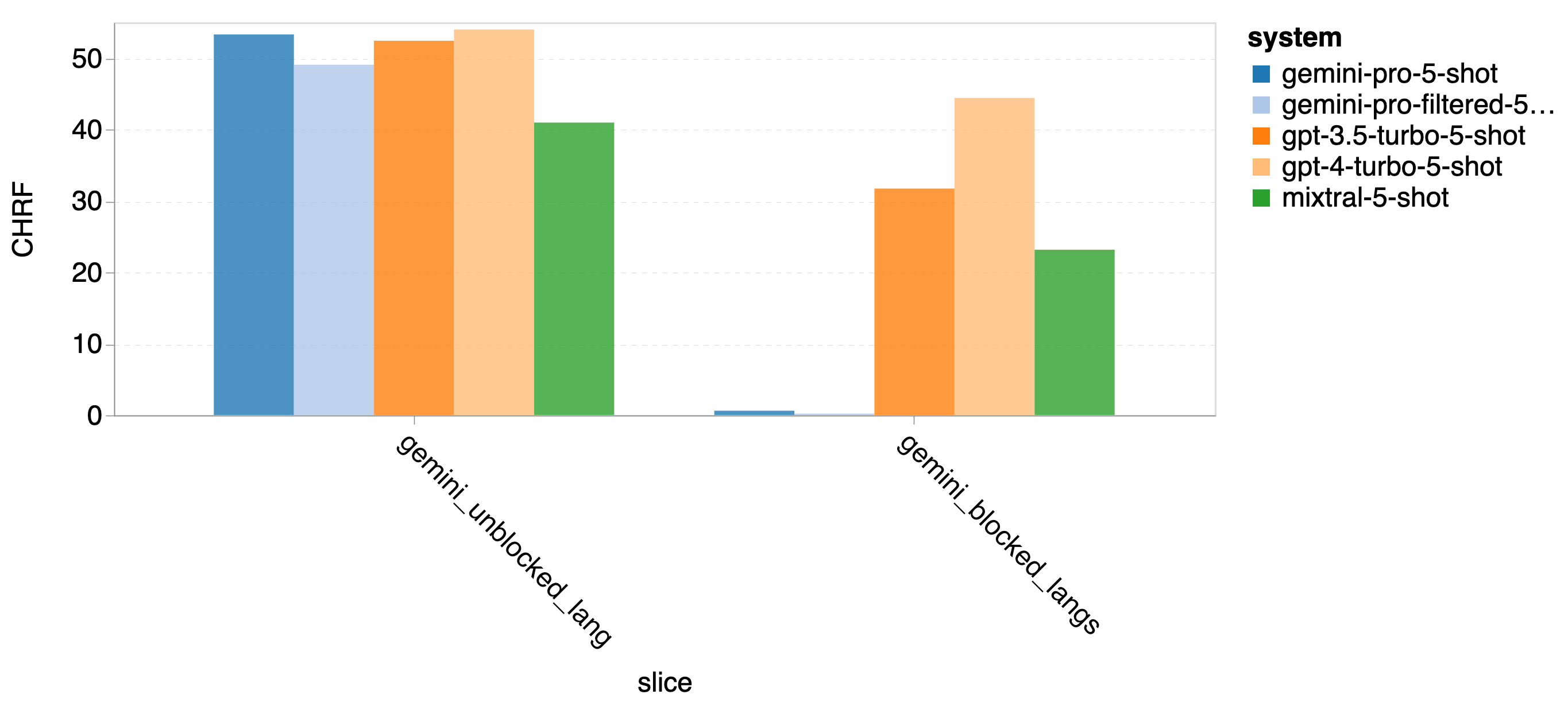}
    \vspace{-3mm}
    \caption{Performance in chrf (\%) on blocked and unblocked languages for 5-shot prompt.}
    \label{fig:mt_blocked_lang_perf_five}
\end{figure}

\paragraph{Gemini Blocked Responses}

If we analyze responses at a language level, we see in \autoref{fig:mt_blocked_five} that Gemini Pro's lower performance in 12/20 languages is due to its tendency to block responses on particular languages, generally ones with lower levels of resources.
We consider a response "blocked" if Gemini Pro generates a \textit{Blocked Response} error, and define unblocked languages as those languages where >50\% samples are not blocked.

Examining performance at a language level in  \autoref{fig:mt_blocked_lang_perf_five}, we see that Gemini Pro outperforms GPT 3.5 Turbo and GPT 4 Turbo on 5/8 unblocked languages most of which are high resource.
Additionally, in Figure \autoref{fig:mt_blocked_lang_perf_five}, we observe that the implementation of safety filters leads to a decrease in the overall chrF score. This reduction occurs because the filters block samples from languages that the model otherwise handles relatively effectively.

\begin{figure}[h!]
\centering
\begin{subfigure}[t]{0.49\textwidth}
    \includegraphics[trim=0 60 0 0, clip, width=\textwidth]{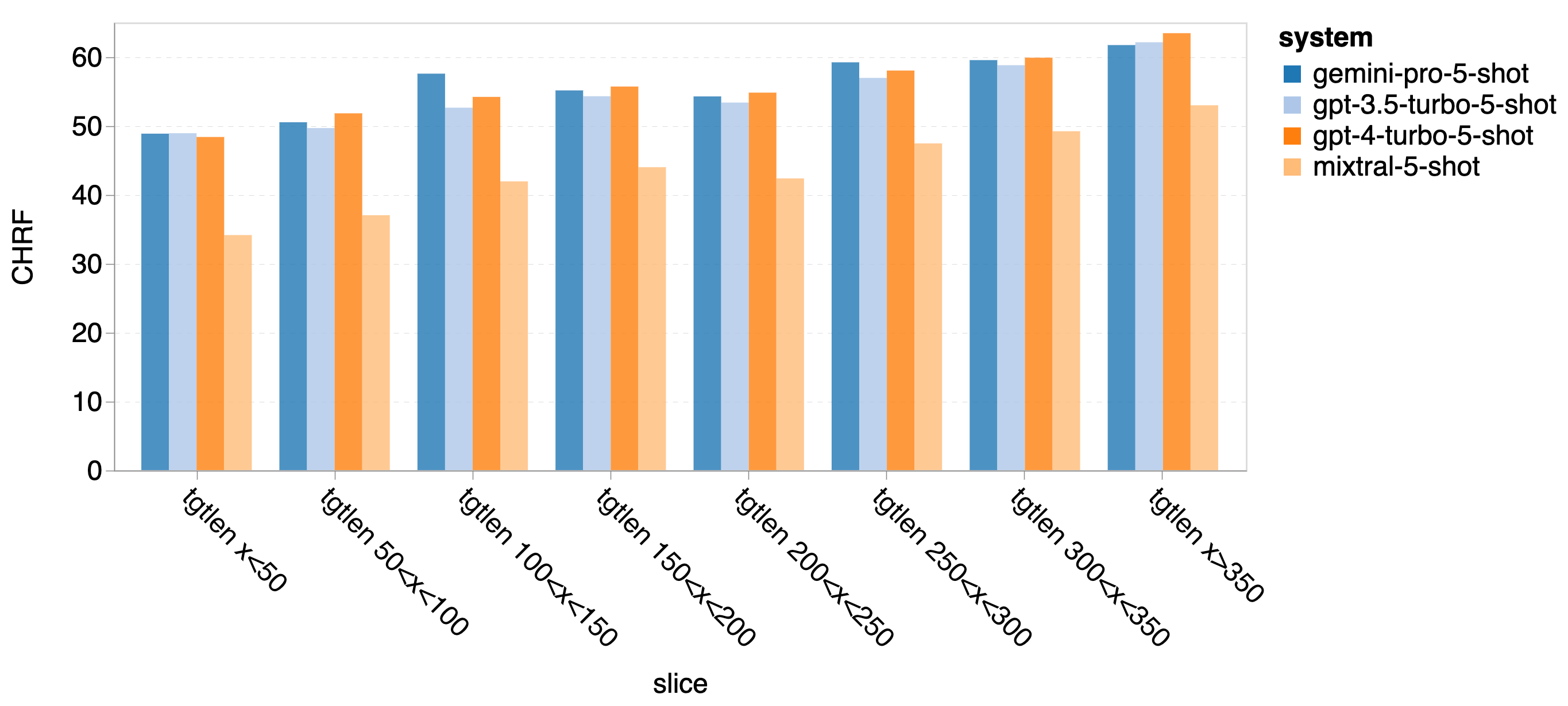}
    \caption{chrf by target sentence length}
    \label{fig:tgtlen}
\end{subfigure}\hspace{\fill} 
\begin{subfigure}[t]{0.49\textwidth}
    \includegraphics[trim=0 60 0 0, clip, width=\linewidth]{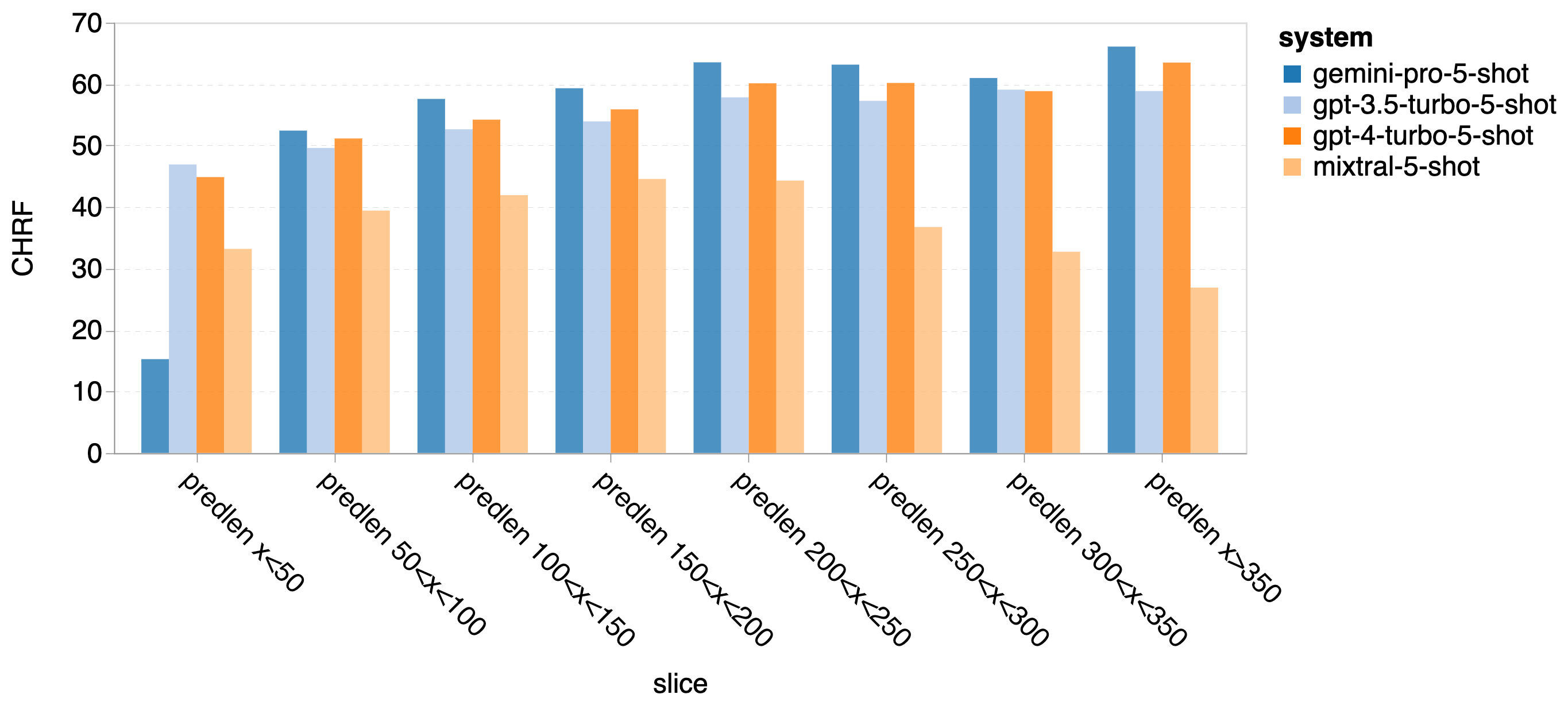}
    \caption{chrf by predicted sentence length}
    \label{fig:mt_predlen}
\end{subfigure}
\caption{Performance with varying target and predicted length using 5-shot prompt on unblocked languages.}
\label{fig:mt_len_perf}
\end{figure}

\paragraph{Other trends}

\begin{wrapfigure}{r}{0.4\textwidth}
    \centering
    \includegraphics[trim=0 0 0 0, clip, width=\textwidth]{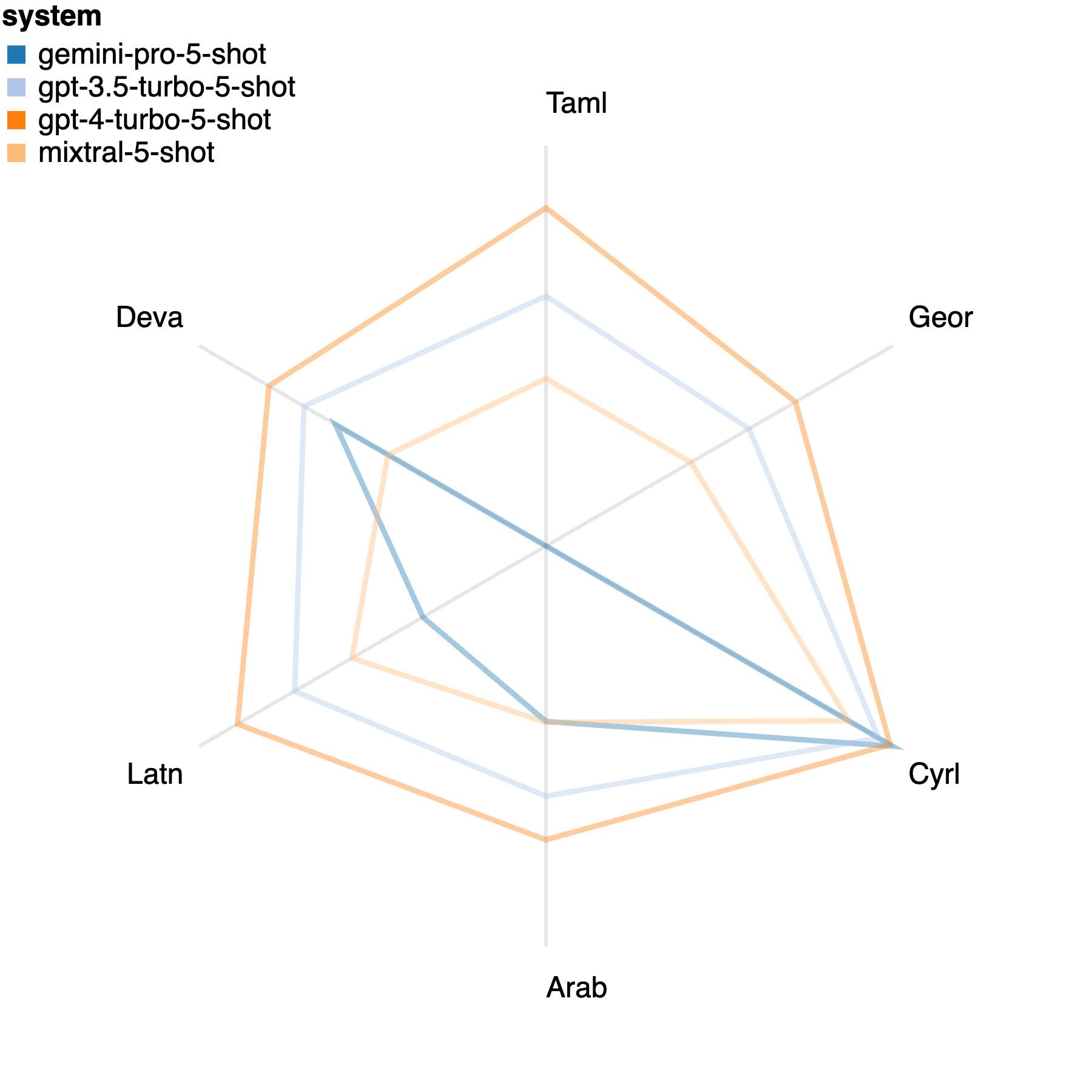}
    \caption{Performance by script}
    \label{fig:mt_script}
     \vspace{-5mm}
\end{wrapfigure}

In \autoref{fig:mt_script}, we present apparent trends when categorizing languages by family or script. A key observation is Gemini Pro's competitive performance with other models on Cyrillic scripts, is contrasted by its underperformance on other scripts. GPT 4 Turbo stands out, outperforming other models across various scripts, especially in the Devanagari script.


\newpage

In Figure \ref{fig:mt_len_perf}, we examine the performance of various models across different sentence length segments, 
categorized by both target length (\autoref{fig:tgtlen}) and predicted length (\autoref{fig:mt_predlen}). Upon scrutinizing Figure \ref{fig:mt_len_perf}, we observe that Gemini Pro's performance at longer target lengths does not match that of GPT 4 Turbo and GPT 3.5 Turbo. However, when considering predicted lengths, Gemini Pro generally outperforms both GPT 4 Turbo and GPT 3.5 Turbo at longer lengths, suggesting it produces higher quality translations at longer lengths. Additionally, the figure suggests that a significant portion of the performance decline at shorter predicted lengths even on unblocked languages may be attributed to empty predictions, likely triggered by the content filtering mechanism.



%% file: tables/mt-5.tex
\begin{table}[t!]
\centering
\begin{tabular}{lrrrrrr}
\toprule
Lang. & Gemini Pro & GPT 3.5 Turbo & GPT 4 Turbo & Mixtral & Google & NLLB \\
\midrule
ssw\_Latin & 0.43$^\dagger$ & 18.16 & \underline{38.62} & 18.84 & - & \textbf{43.51} \\
sna\_Latin & 0.00$^\dagger$ & 24.44 & 43.25 & 22.20 & \textbf{44.36} & \underline{43.40} \\
ckb\_Arab & 0.00$^\dagger$ & 26.69 & 41.30 & 18.36 & \textbf{47.61} & \underline{47.25} \\
mag\_Deva & 34.54 & 39.70 & \underline{45.46} & 25.93 & - & \textbf{58.03} \\
ibo\_Latin & 0.00$^\dagger$ & 21.46 & \underline{41.94} & 17.75 & \textbf{43.37} & 41.36 \\
hau\_Latin & 0.02$^\dagger$ & 30.24 & 50.82 & 22.47 & \underline{53.18} & \textbf{53.25} \\
pbt\_Arab & 0.00$^\dagger$ & 22.81 & \underline{34.21} & 16.61 & - & \textbf{39.27} \\
tam\_Taml & 0.00$^\dagger$ & 35.50 & 48.04 & 23.78 & \textbf{55.98} & \underline{53.63} \\
kat\_Geor & 0.00$^\dagger$ & 33.32 & 40.94 & 23.78 & \textbf{51.11} & \underline{47.10} \\
gle\_Latin & 0.00$^\dagger$ & 46.72 & 56.52 & 26.93 & \textbf{59.93} & \underline{57.87} \\
kmr\_Latin & 0.00$^\dagger$ & 30.03 & 33.33 & 19.04 & \textbf{39.94} & \underline{39.25} \\
war\_Latin & 0.52$^\dagger$ & 51.17 & \underline{56.01} & 34.74 & - & \textbf{57.25} \\
ajp\_Arab & \textbf{50.64} & 47.45 & 47.01 & 33.60 & - & \underline{50.61} \\
lim\_Latin & 39.99 & 43.77 & \underline{46.05} & 32.31 & - & \textbf{47.58} \\
ukr\_Cyrl & \underline{56.89} & 54.56 & 56.44 & 49.66 & \textbf{58.25} & 56.04 \\
fra\_Latin & 70.77 & \underline{70.99} & 70.77 & 66.73 & \textbf{72.98} & 70.01 \\
lvs\_Latin & \underline{59.49} & 54.55 & 57.95 & 31.17 & \textbf{62.49} & 54.89 \\
ron\_Latin & \textbf{65.09} & 63.18 & 63.93 & 56.68 & \underline{65.08} & 61.21 \\
tpi\_Latin & 6.20$^\dagger$ & 40.14 & \textbf{47.97} & 33.33 & - & \underline{42.02} \\
acm\_Arab & \textbf{49.05} & 45.26 & \underline{44.44} & 31.65 & - & 32.60 \\
\bottomrule
\end{tabular}
\caption{Machine translation performance (chrF (\%) scores) across models for all languages using 5-shot prompt. Best scores are bolded, second best underlined. $^\dagger$ indicates languages where Gemini Pro blocked more than 50\% of responses.}
\label{table:mt2}
\end{table}

%% file: sections/08_WebAgent.tex
\resultsection{Web Agents}{https://hub.zenoml.com/report/2608/Gemini\%20Webarena}
\label{sec:webarena}

\input{tables/WebArena}

Finally, we examine the ability of each model to act as an instruction following agent that performs tasks on the web, which requires long-term planning and complex data understanding.
We use \texttt{WebArena} \citep{zhou2023webarena}, an execution-based simulation environment where the success criterion is based on execution outcome.
Tasks given to agents consist of information
seeking, site navigation, and content \& configuration operations.
The tasks span over a variety of web sites, including E-commerce platforms, social forums, collaborative software development platforms (e.g.~gitlab), content management systems, and online maps.

\subsection{Experiment Details}

\paragraph{Generation Parameters}

We follow \texttt{WebArena}'s testing methodology in testing Gemini. We used the two-shot chain-of-thought prompts from  \citet{zhou2023webarena}, where each prompt includes two CoT style examples.
We further distinguished between whether or not the model is instructed to terminate execution when it believes the task is unachievable (the ``unachievable'' hint, or \texttt{UA} in \texttt{WebArena} parlance).

In sum, we tested with two prompts from \texttt{WebArena}: \texttt{p\_cot\_id\_actree\_2s} and \texttt{p\_cot\_id\_actree\_2s\_no\_na}, which are the CoT prompt with the \texttt{UA} hint and CoT prompt without the \texttt{UA} hint, respectively. To make results comparable between GPTs and Gemini, we set the same upper limit on the observation lengths for all of them. This number is set to 1920 tokens using the tokenizer of \texttt{gpt-4-1106-preview}, consistent with experiments in \texttt{WebArena}. In terms of hyper-parameters, we used the default suggested by each of the large language model providers. For the Gemini models, the suggested default temperature is 0.9 and default top-p is 1.0, and the \texttt{WebArena} suggested default for GPT models is 1.0 for temperature and 0.9 for top-p.

\paragraph{Evaluation Procedure}
The action sequence of an agent is considered correct as long as they achieved the final goal, regardless the intermediate steps they take. 
We use \texttt{WebArena}'s evaluation, which determines wether a task is completed successfully or not with the agent's final output.

\begin{wrapfigure}{r}{0.35\textwidth}
    \centering
    \vspace{-10mm}
    \includegraphics[width=1.0\textwidth]{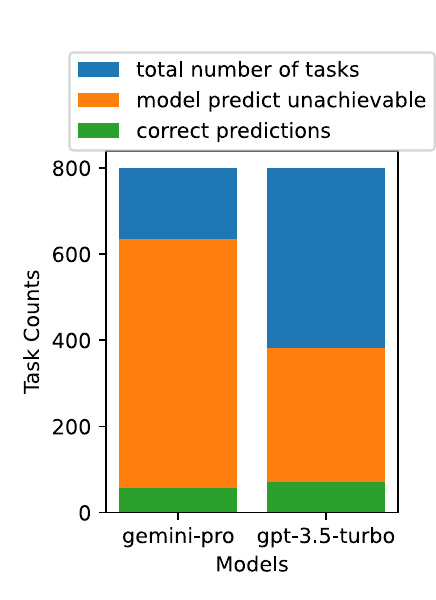}
    \vspace{-5mm}
    \caption{UA prediction count\label{fig:WebArena-ua}}
\end{wrapfigure}

\subsection{Results and Analysis}

We examine Gemini-Pro's overall success rate, rate across different tasks, its response lengths, trajectory step counts, and tendency to predict that the task is unachievable.
The overall performance is list in \autoref{tab:webarena-result}. Gemini-Pro performs comparably but slightly worse than GPT-3.5-Turbo.
Similarly to GPT-3.5-Turbo, Gemini-Pro performs better when the prompt mentions that task might be unachievable (\texttt{UA hint}). With \texttt{UA} hint, Gemini-Pro achieves an overall 7.09 percent success rate.

\begin{figure}[t]
    \centering
    \includegraphics[trim=0 18 0 0, clip, width=1.0\textwidth]{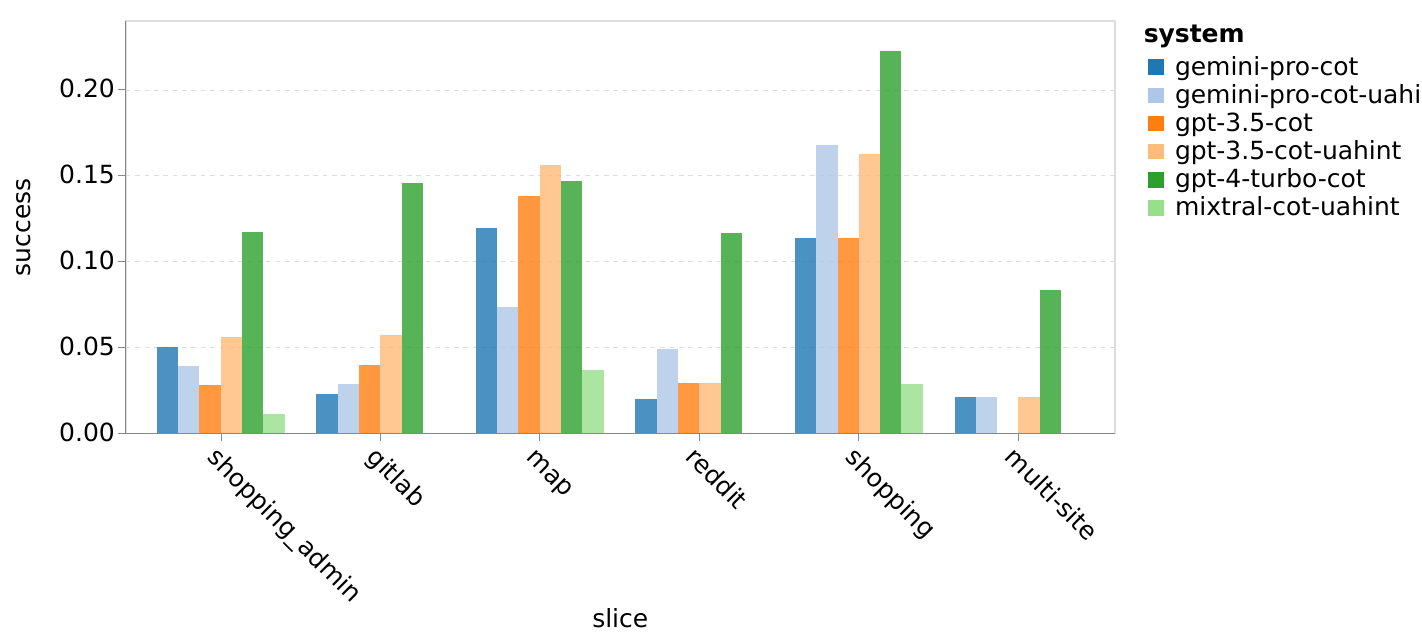}
    \caption{Web agent success rate of evaluated models at different site groups\label{fig:WebArena-site}}
\end{figure}

If we break down by websites, as shown in \autoref{fig:WebArena-site}, we can see that Gemini-Pro performs worse than GPT-3.5-Turbo on gitlab and maps, while being close to GPT-3.5-Turbo on shopping admin, reddit, and shopping.
It performs better than GPT-3.5-Turbo on multi-site tasks, which is in concert with our previous results of Gemini being a bit better on the more complex sub-tasks across benchmarks.

In general, Gemini-Pro predicts more tasks as unachievable, especially in the case where a \texttt{UA} hint is given, as shown in \autoref{fig:WebArena-ua}. Gemini-Pro predicts over 80.6\% of the tasks as unachievable when given a \texttt{UA} hint, compared to 47.7\% by GPT-3.5-Turbo. Note that 4.4\% of the tasks in the dataset are actually unachievable, so both far over-predict the actual number of unachievable tasks.

\begin{figure}[ht]
\begin{subfigure}[t]{0.47\textwidth}
    \centering
    \includegraphics[trim=0 9.5 0 0, clip, width=1\textwidth]{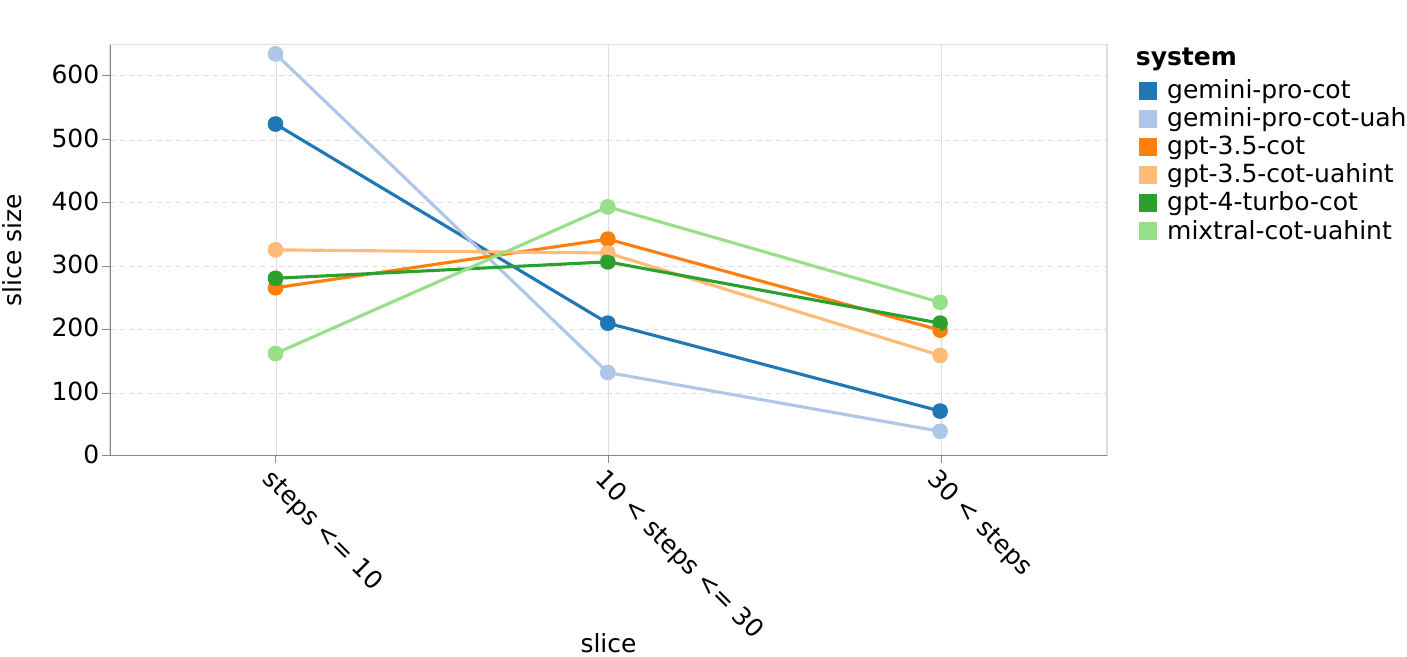}
    \caption{Average steps taken per task \label{fig:WebArena-steps}}
\end{subfigure}
\begin{subfigure}[t]{0.47\textwidth}
    \centering
    \includegraphics[trim=0 8.5 0 0, clip, width=1\textwidth]{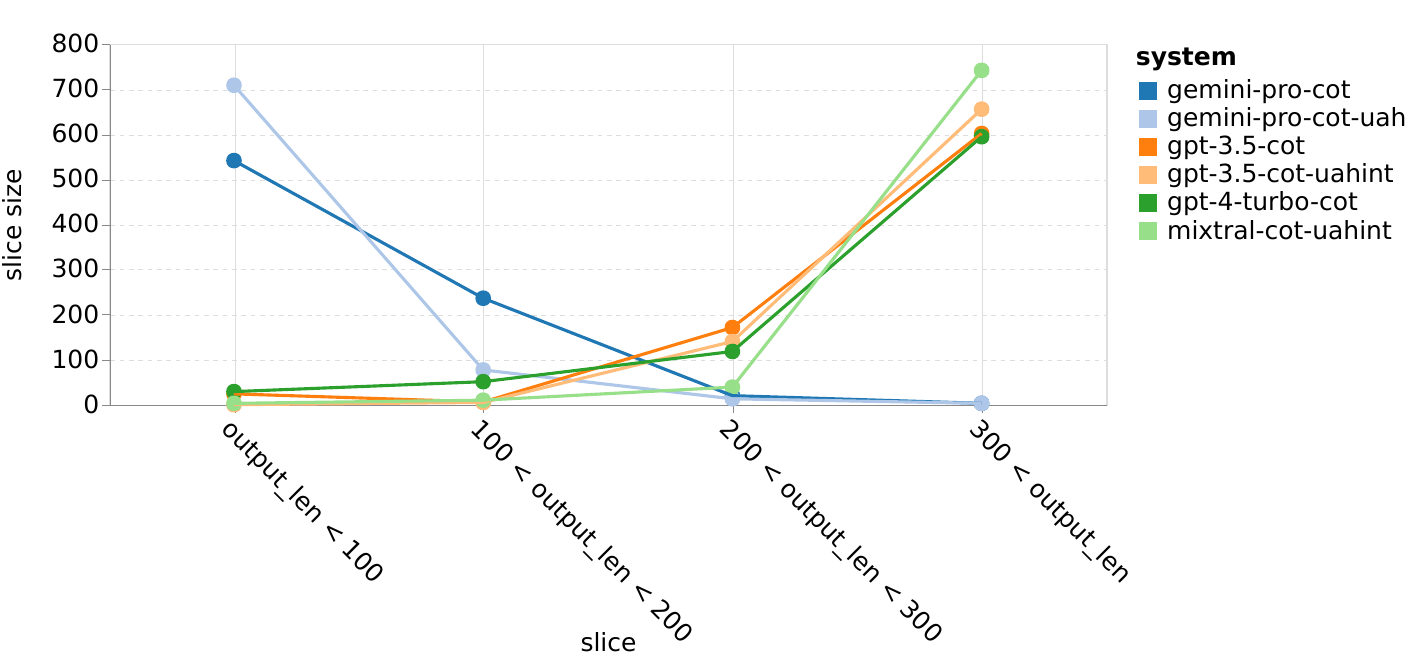}
    \caption{Average response length \label{fig:WebArena-len}}
\end{subfigure}
\caption{Model behaviors on \texttt{WebArena}.}
\end{figure}

At the same time, we observed that Gemini Pro has a greater tendency to respond in shorter phrases and take fewer steps before reaching a conclusion. As shown in \autoref{fig:WebArena-steps}, more than half of trajectories by Gemini Pro are under ten steps, while majority of trajectories by GPT 3.5 Turbo and GPT 4 Turbo are between 10 and 30 steps. Similarly, the majority of Gemini responses are less than 100 characters in length, while most of GPT 3.5 Turbo, GPT 4 Turbo, and Mixtral's responses are over 300 characters in length \autoref{fig:WebArena-len}. Gemini tends to directly predict the actions while other models would start with reasoning and then give their action predictions.

%% file: tables/webarena.tex
\begin{wraptable}{r}{0pt}
    \begin{tabular}{@{}c@{\hspace{13pt}}c@{}l@{\hspace{15pt}}ccc@{}}
        \toprule
        \textbf{CoT} & \textbf{UA Hint} & \multicolumn{1}{c}{\textbf{Model}} & \textbf{SR} & \textbf{SR$_\textrm{AC}$}\\
        \midrule
        \cmark & \cmark & \scriptsize \textsc{Gemini-pro} & 7.12 & 3.52 \\
        \cmark & \xmark & \scriptsize\textsc{Gemini-pro} &  6.25 & 4.83 \\
        \cmark & \cmark & \scriptsize \textsc{GPT-3.5-turbo} & 8.87 & 6.44  \\
        \cmark & \xmark & \scriptsize\textsc{GPT-3.5-turbo} & 6.36 & 6.06  \\
        \cmark & \xmark & \scriptsize\textsc{GPT-4-turbo} & \textbf{14.90} & \textbf{14.22} \\ 
        \bottomrule
    \end{tabular}
    \caption{Performances on \texttt{WebArena}.}
    \label{tab:webarena-result}
\end{wraptable}

%% file: sections/09_Conclusion.tex
\section{Conclusion}

In this paper, we have taken a first \emph{impartial, in-depth} look into Google's Gemini model, comparing it to OpenAI's GPT 3.5 and 4 models, as well as the open source Mixtral model.

\paragraph{Takeaways}
We came away with a number of conclusions:
\begin{itemize}
\item The Gemini Pro model, which is comparable to GPT 3.5 Turbo in model size and class, generally achieves accuracy that is comparable but somewhat inferior to GPT 3.5 Turbo, and much worse than GPT 4 on English tasks.
\item In particular, we find that Gemini Pro was somewhat less performant than GPT 3.5 Turbo on average, but in particular had issues of bias to response order in multiple-choice questions, mathematical reasoning with large digits, and premature termination of agentive tasks. When using the default content filtering settings, there were also failed responses due to aggressive content filtering.
\item On the other hand, there were bright points: Gemini performed better than GPT 3.5 Turbo on particularly long and complex reasoning tasks.
\item In addition, when generating text in other languages (specifically through translation), Gemini Pro outperforms both GPT 3.5 Turbo and GPT 4 Turbo on the languages where requests are not blocked, but there are several languages for which Gemini Pro does not return any answer.
\item The open-source model Mixtral is competitive with Gemini Pro and GPT 3.5 Turbo on Knowledge-based QA and Mathematics tasks, but significantly underperformed on other tasks. 
\end{itemize}

\paragraph{Limitations}
Finally, we would like to temper these conclusions with a number of limitations.

First, our work is a snapshot in time with respect to ever-changing and unstable API-based systems.
All results here are current as of this writing on \today, but may change in the future as models and the surrounding systems are upgraded.

Second, the results may be dependent on the specific prompts and generation parameters that we selected.
In fact, we found that the results of all models were affected by prompt selection, and that the GPT models seemed somewhat more robust to small variations in the prompts of the GPT models.
It is quite possible that with further prompt engineering, or multiple samples and self-consistency as was used by \citet{gemini23gemini}, the results could change significantly.

Finally, any benchmarking paper would be remiss without a discussion of data leakage, which plagues current evaluation of large language models \citep{zhou2023don}.
While we did not measure this leakage explicitly, we did attempt to mitigate by evaluating on a broad variety of tasks, including those who's outputs were not sourced from or widely available on the internet (such as \texttt{WebArena}).

\paragraph{Outlook}
Based on this paper, we can make the recommendation to researchers and practitioners to carefully look at the Gemini Pro model as a tool in the toolbox, comparable to GPT 3.5 Turbo.
In particular, Gemini Pro may be a preferable alternative when processing non-English languages.
Gemini's Ultra edition, which is yet to be released, is reported to be on par with GPT 4, and a further examination of this model will be warranted when it is available.

%% file: sections/A01_contributions.tex
\section{Author Contributions}
\label{sec:author_contributions}

Syeda Akter performed experiments, analysis, and writing for the text understanding and mathematical reasoning tasks.
Zichun Yu performed experiments, analysis, and writing for the knowledge-based question answering and the code generation tasks.
Aashiq Muhamed performed experiments, analysis, and writing for the machine translation task.
Tianyue Ou performed experiments, analysis, and writing for the instruction following agents task.
Ángel Alexander Cabrera and Alex Bäuerle provided visualization support and performed fine-grained analysis of the results for each tasks.
Krrish Dholakia provided support implementing calls to each of the language models.
Chenyan Xiong provided direction on the varieties of tasks to pursue and helped with paper writing.
Graham Neubig proposed the project idea, wrote the introduction, experimental setup, and conclusions section, and provided analysis and writing support for all other sections.

%% file: sections/A02_prompt_details.tex
\section{Prompt Details}
\label{sec:prompt_details}

In this section, we detail the prompts that we used for each task.

For Knowledge-based QA task in Section \ref{sec:mmlu}, we have used standard 5-shot prompts from \cite{hendrycks2020measuring}\footnote{\url{https://github.com/hendrycks/test}} and 5-shot chain-of-thought prompts from chain-of-thought-hub\footnote{\url{https://github.com/FranxYao/chain-of-thought-hub/blob/main/MMLU/lib_prompt/mmlu-cot.json}}.

For General-purpose Reasoning task in Section \ref{sec:bbh}, we have used Chain-of-Thought prompts from \cite{eval-harness}\footnote{\url{https://github.com/EleutherAI/lm-evaluation-harness/tree/big-refactor/lm_eval/tasks/bbh/cot_fewshot}}.

For Mathematics tasks in Section \ref{sec:math}, we also have followed Chain-of-Thought prompts from \cite{eval-harness}\footnote{\url{https://github.com/EleutherAI/lm-evaluation-harness/blob/big-refactor/lm_eval/tasks/gsm8k/gsm8k-cot.yaml}}.

For Code Generation in Section \ref{sec:code}, prompt is listed in \autoref{tab:code_gen_prompts}.

\input{tables/code_prompt}

For Machine Translation in Section \ref{sec:mt}, prompts are listed in \autoref{tab:translation_prompts}.

\begin{table}[htbp]
\centering
\begin{tabular}{@{}cl@{}}
\toprule
\textbf{Shot} & \textbf{Prompt} \\
\midrule
zero & \begin{tabular}[c]{@{}l@{}}This is an English to [TGT] translation, please provide \\ the [TGT] translation for this sentence. Do not provide \\ any explanations or text apart from the translation. \\ {[}SRC{]}: [src-sentence] \\ {[}TGT{]}: \end{tabular} \\
\addlinespace
five & \begin{tabular}[c]{@{}l@{}}This is an English to [TGT] translation, please provide \\ the [TGT] translation for these sentences: \\ {[}SRC{]}: [src-sentence] {[}TGT{]}: [tgt-sentence] \\ {[}SRC{]}: [src-sentence] {[}TGT{]}: [tgt-sentence] \\ {[}SRC{]}: [src-sentence] {[}TGT{]}: [tgt-sentence] \\ {[}SRC{]}: [src-sentence] {[}TGT{]}: [tgt-sentence] \\ {[}SRC{]}: [src-sentence] {[}TGT{]}: [tgt-sentence] \\ Please provide the translation for the following sentence. \\ Do not provide any explanations or text apart from the \\ translation. \\ {[}SRC{]}: [src-sentence] \\ {[}TGT{]}: \end{tabular} \\
\bottomrule
\end{tabular}
\captionsetup{skip=10pt} 
\caption{Prompts used for zero- and five-shot settings in translation tasks.}
\label{tab:translation_prompts}
\end{table}

For WebArena in Section \ref{sec:webarena}, we used \href{https://github.com/oootttyyy/webarena/blob/main/agent/prompts/raw/p_cot_id_actree_2s.py}{CoT with \texttt{UA} (unachievable) hint}\footnote{\url{https://github.com/oootttyyy/webarena/blob/main/agent/prompts/raw/p\_cot\_id\_actree\_2s.py}} and \href{https://github.com/oootttyyy/webarena/blob/main/agent/prompts/raw/p_cot_id_actree_2s_no_na.py}{CoT without \texttt{UA} hint}\footnote{\url{https://github.com/oootttyyy/webarena/blob/main/agent/prompts/raw/p\_cot\_id\_actree\_2s\_no\_na.py}}. 

%% file: tables/code_prompt.tex
\begin{table}[htbp]
\centering
\begin{tabular}{l}
\toprule
\textbf{Prompt} \\
\midrule
Write the following python3 function:  \\
$[$CODE BLOCK$]$\\
\addlinespace
\bottomrule
\end{tabular}
\captionsetup{skip=10pt} 
\caption{Prompts used for code generation tasks.}
\label{tab:code_gen_prompts}
\end{table}

%% file: sections/A03_mt_additional.tex
\section{Additional Experiments: Machine Translation}
\label{sec:mt_addn_exps}

This section includes plots and results comparing 0-shot and 5-shot prompts for all models, as well as the results using 0-shot prompt per language.
Throughout our analysis of the models, we observed that few-shot prompts generally yield a modest enhancement in average performance, with an increasing variance pattern following the order: GPT 3.5 Turbo < Gemini Pro < GPT 4 Turbo < Mixtral. Using 5-shot prompts improves chrf for both unblocked language instances as well as all instances. Gemini Pro gains 1.8 chrf, GPT 5 Turbo gains 2.87 chrf, GPT 2.5 Turbo gains 0.8 chrf, and Mixtral gains 4.75 chrf.

\input{tables/mt-0}

\begin{figure}[ht!]
    \centering
    \includegraphics[width=\textwidth]{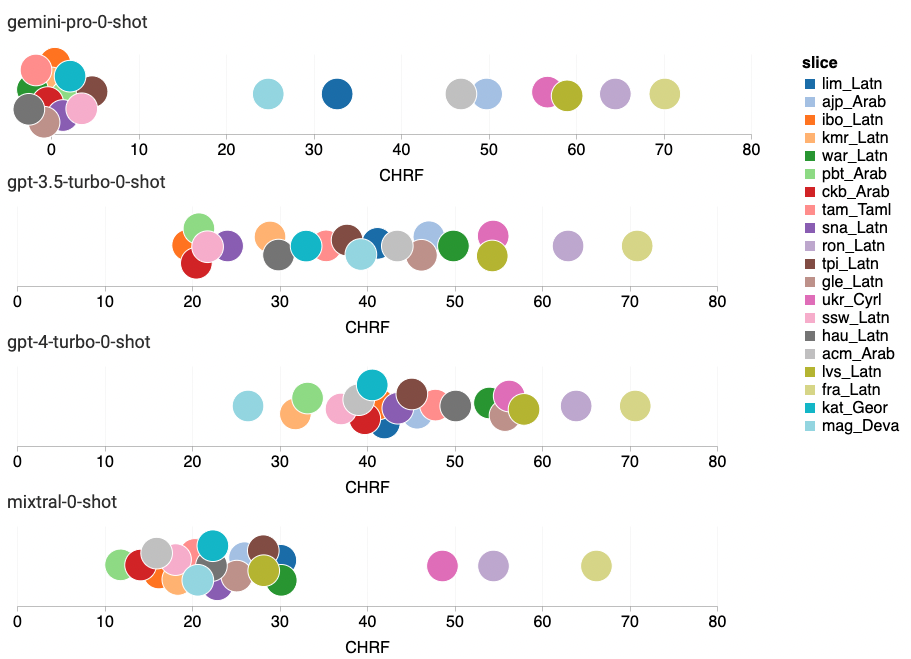}
    \caption{Machine translation performance (chRF (\%) scores) by language pairs for 0-shot prompt.}
    \label{fig:mt_bubble_zero}
\end{figure}

\begin{figure}[h!]
    \centering
    \includegraphics[trim=0 20 0 0, clip, width=0.95\textwidth]{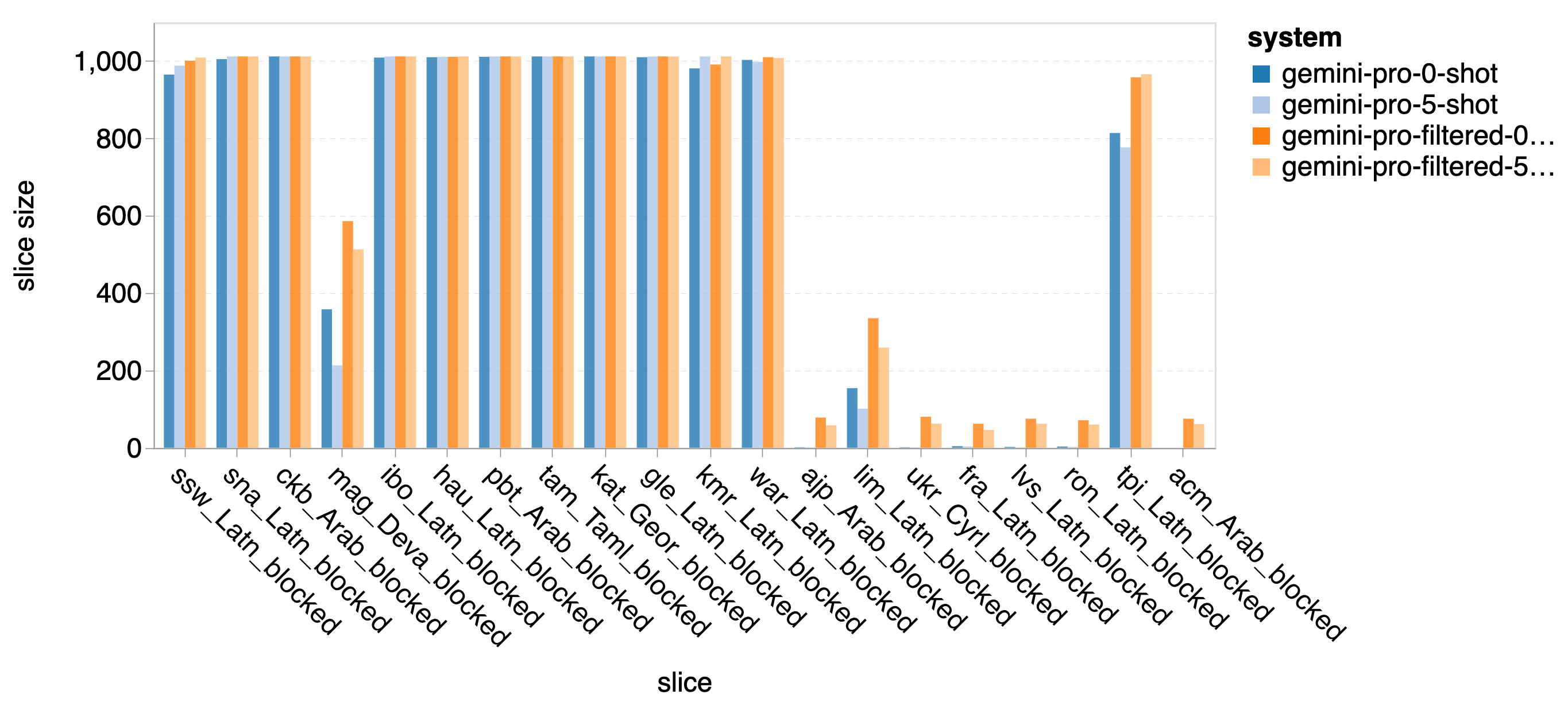}
    \vspace{-3mm}
    \caption{Number of samples that are blocked by Gemini Pro for 0-shot prompt and 5-shot prompt.}
    \label{fig:mt_blocked_zero}
\end{figure}


\begin{figure}[h!]
    \centering
    \includegraphics[trim=0 15 0 0, clip, width=0.95\textwidth]{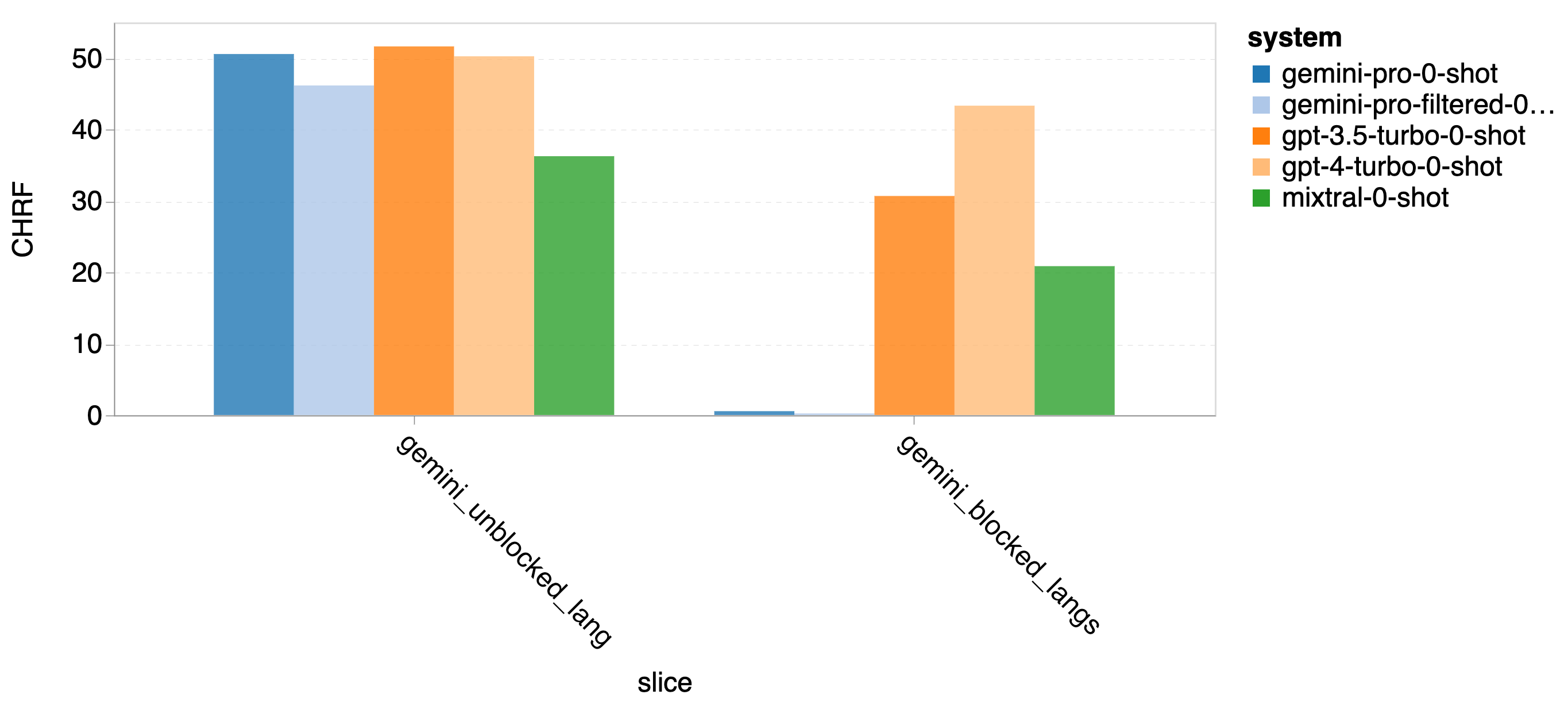}
    \vspace{-3mm}
    \caption{Performance in chrf (\%) on blocked and unblocked languages for 0-shot prompt}
    \label{fig:mt_blocked_lang_perf_zero}
\end{figure}

%% file: tables/mt-0.tex
\begin{table}[t!]
\centering
\begin{tabular}{lrrrrrrr}
\toprule
Lang. & Gemini Pro & GPT 3.5 Turbo & GPT 4 Turbo & Mixtral & Google & NLLB \\
\midrule
ssw\_Latin & 0.54$^\dagger$ & 20.21 & \underline{37.38} & 18.27  & - & \textbf{43.51} \\
sna\_Latin & 0.11$^\dagger$ & 23.92 & 42.84 & 22.07 & \textbf{44.36} & \underline{43.40} \\
ckb\_Arab & 0.00$^\dagger$ & 21.13 & 39.90 & 14.31  & \textbf{47.61} & \underline{47.25} \\
mag\_Deva & 24.83 & \underline{39.20} & 26.40 & 20.29 & - & \textbf{58.03} \\
ibo\_Latin & 0.08$^\dagger$ & 20.13 & \underline{41.78} & 16.71 & \textbf{43.37} & 41.36 \\
hau\_Latin & 0.05$^\dagger$ & 29.64 & 50.14 & 22.41 & \underline{53.18} & \textbf{53.25} \\
pbt\_Arab & 0.01$^\dagger$ & 21.26 & \underline{32.85} & 12.46  & - & \textbf{39.27} \\
tam\_Taml & 0.00$^\dagger$ & 35.13 & 47.67 & 20.97 & \textbf{55.98} & \underline{53.63} \\
kat\_Geor & 0.00$^\dagger$ & 33.24 & 40.45 & 21.42 & \textbf{51.11} & \underline{47.10} \\
gle\_Latin & 0.09$^\dagger$ & 46.44 & 55.91 & 25.34 & \textbf{59.93} & \underline{57.87} \\
kmr\_Latin & 0.57$^\dagger$ & 29.16 & 32.17 & 17.96 & \textbf{39.94} & \underline{39.25} \\
war\_Latin & 0.38$^\dagger$ & 49.88 & \underline{54.21} & 29.66 & - & \textbf{57.25} \\
ajp\_Arab & \underline{49.79} & 46.86 & 45.32 & 26.07 & - & \textbf{50.61} \\
lim\_Latin & 32.70 & 41.00 & \underline{42.04} & 29.13 & - & \textbf{47.58} \\
ukr\_Cyrl & \underline{57.20} & 54.39 & 56.51 & 48.62 & \textbf{58.25} & 56.04 \\
fra\_Latin & \underline{70.15} & 70.88 & 70.66 & 66.22 & \textbf{72.98} & 70.01 \\
lvs\_Latin & \underline{58.53} & 54.34 & 57.34 & 28.99 & \textbf{62.49} & 54.89 \\
ron\_Latin & \textbf{64.50} & 62.98 & 63.90 & 54.46 & \underline{65.08} & 61.21 \\
tpi\_Latin & 4.62$^\dagger$ & 38.05 & \textbf{44.70} & 28.63 & - & \underline{42.02} \\
acm\_Arab & \textbf{46.85} & \underline{43.44} & 39.83 & 16.26  & - & 32.60 \\
\bottomrule
\end{tabular}
\caption{Machine translation performance (chrF (\%) scores) across models for all languages using 0-shot prompt. Best scores are bolded, second best underlined. $^\dagger$ indicates languages where Gemini Pro blocked more than 50\% of responses.}
\label{table:mt1}
\end{table}